\pdfoutput=1
\PassOptionsToPackage{table}{xcolor}
\documentclass[11pt]{article}

\usepackage[]{acl}
\usepackage{times}
\usepackage{latexsym}
\usepackage{booktabs}
\usepackage{amsmath}
\usepackage{multirow}
\usepackage{tabularx}
\usepackage{caption}
\usepackage{rotating}
\usepackage{graphicx}

\usepackage{pgfplotstable}
\usepackage{xcolor,colortbl}
\usepackage{CJKutf8}
\usepackage{wasysym}
\usepackage{tikz}
\usepackage[table]{xcolor}
\usepackage{cleveref}
\crefformat{section}{\S#2#1#3} 
\crefformat{subsection}{\S#2#1#3}
\crefformat{subsubsection}{\S#2#1#3}

\definecolor{ca}{RGB}{240, 116, 112}
\definecolor{cb}{RGB}{240, 135, 142}
\definecolor{cc}{RGB}{241, 149, 155}
\definecolor{cd}{RGB}{243, 160, 165}
\definecolor{ce}{RGB}{246, 189, 192}
\definecolor{cf}{RGB}{248, 200, 204}

\definecolor{cg}{RGB}{255,255,255}
\definecolor{ch}{RGB}{255,255,255}
\definecolor{ci}{RGB}{255,255,255}
\definecolor{cj}{RGB}{255,255,255}
\definecolor{ck}{RGB}{255,255,255}
\definecolor{cl}{RGB}{255,255,255}

\definecolor{red}{HTML}{eb6c69}
\definecolor{green}{HTML}{81d653}
\definecolor{orange}{HTML}{FDB594}
\definecolor{blue_}{HTML}{B8D2E3}
\newcommand{\greenone}{\cellcolor[HTML]{98C3DD}}
\newcommand{\greentwo}{\cellcolor[HTML]{B8D2E3}}
\newcommand{\greenthree}{\cellcolor[HTML]{CEDDE7}}
\newcommand{\redone}{\cellcolor[HTML]{FC8D59}}
\newcommand{\redtwo}{\cellcolor[HTML]{FDB594}}
\newcommand{\redthree}{\cellcolor[HTML]{FDC7AE}}

\usepackage[T1]{fontenc}

\usepackage[utf8]{inputenc}
\usepackage{amsmath}
\DeclareMathOperator*{\argmax}{arg\,max}

\DeclareMathOperator*{\argtopk}{argtopk}
\DeclareMathOperator*{\score}{score}
\usepackage{microtype}
\usepackage{graphicx}
\usepackage{inconsolata}
\usepackage{float}
\usepackage{enumitem}

\title{A Thorough Examination of Decoding Methods in the Era of LLMs}

\author{Chufan Shi$^{\spadesuit}$\Thanks{\hspace{1mm}Equal Contribution. Code is available at~\url{https://github.com/DavidFanzz/llm_decoding.git}}\hspace{1.0mm}, 
    Haoran Yang$^{\clubsuit*}$\hspace{0.5mm},
    Deng Cai$^{\heartsuit}$\Thanks{\hspace{1mm}Corresponding authors.}\hspace{0.5mm}, \\
    \textbf{Zhisong Zhang}$^{\heartsuit}$\hspace{0.5mm}, 
    \textbf{Yifan Wang}$^{\spadesuit}$\hspace{0.5mm}, 
    \textbf{Yujiu Yang}$^{\spadesuit\dag}$\hspace{0.5mm}, 
    \textbf{Wai Lam}$^{\clubsuit}$\hspace{0.5mm} \\
    $^{\spadesuit}$Tsinghua University~$^{\clubsuit}$The Chinese University of Hong Kong~$^{\heartsuit}$Tencent AI Lab\\
    \texttt{\{scf22,wangyifa22\}@mails.tsinghua.edu.cn \{hryang,wlam\}@se.cuhk.edu.hk} \\
    \texttt{\{jcykcai, zhisonzhang\}@tencent.com yang.yujiu@sz.tsinghua.edu.cn} \\}
\begin{document}
\maketitle
\begin{abstract}
Decoding methods play an indispensable role in converting language models from next-token predictors into practical task solvers. Prior research on decoding methods, primarily focusing on task-specific models, may not extend to the current era of general-purpose large language models (LLMs). Moreover, the recent influx of decoding strategies has further complicated this landscape.
This paper provides a comprehensive and multifaceted analysis of various decoding methods within the context of LLMs, evaluating their performance, robustness to hyperparameter changes, and decoding speeds across a wide range of tasks, models, and deployment environments. Our findings reveal that decoding method performance is notably task-dependent and influenced by factors such as alignment, model size, and quantization. Intriguingly, sensitivity analysis exposes that certain methods achieve superior performance at the cost of extensive hyperparameter tuning, highlighting the trade-off between attaining optimal results and the practicality of implementation in varying contexts.

\end{abstract}
\section{Introduction}
The advent of large language models (LLMs)~\cite[][\textit{inter alia}]{2022OpenAIchatgpt,2023GPT4Openai,touvron2023llama,touvron2023llama2} has ushered in a new era of natural language processing (NLP). These models are trained to predict the next token on massive corpora, empowering them with extraordinary multitasking capabilities. This enables them to perform almost all NLP tasks through the lens of text generation, distinguishing them from traditional task-specific models.

Decoding methods, which are the bridge between \textit{next-token predictors} and \textit{text generators}, play an integral role in transforming LLMs into practical task solvers. Recent studies have shown that the choice of decoding methods can substantially impact the performance of LLMs \cite{o2023contrastive,chuang2023dola}. However, these studies often focus on a narrow aspect (e.g., factuality~\cite{chuang2023dola}) and a limited set of similar tasks (e.g., math problem solving~\cite{li2022contrastive}). Notably, \citet{ippolito-etal-2019-comparison,wiher2022decoding} provide a comparative analysis of various decoding methods using task-specific language models. They find that deterministic decoding methods (e.g., beam search) perform better than stochastic decoding methods (e.g., top-$p$ sampling~\cite{Holtzman2020The}) in closed-ended generation tasks such as machine translation, while the inverse is true for open-ended generation tasks such as story generation. However, their findings are confined to traditional task-specific models prior to the advent of LLMs. It is uncertain whether their conclusions still hold for general-purpose LLMs. In addition, a plethora of new decoding methods \cite{su2022a,li2022contrastive,yang2023frustratingly,meister-etal-2023-locally,hewitt2022truncation,basu2021mirostat} have been proposed afterward, each claiming to outperform the previous state-of-the-art in particular tasks. Nevertheless, today's most performant LLMs such as ChatGPT and GPT4 \cite{2022OpenAIchatgpt,2023GPT4Openai} only provide APIs for temperature and top-$p$ sampling, seemingly overlooking the potential benefits of other advanced decoding methods.

The above observations raise a natural question: what is the best practice for choosing decoding methods in the era of LLMs? A thorough analysis of decoding methods is essential for researchers and practitioners to understand the strengths and weaknesses of different decoding methods and to choose the one that best fits their needs. Our work fills this gap by providing a comprehensive study of the \textit{performance}, \textit{robustness}, and \textit{speed} of various decoding methods across a wide range of different tasks, models, and deployment environments. We also provide in-depth analyses to uncover the underlying reasons for the observed results. Our key findings include the following:
\begin{itemize}[wide=0\parindent,noitemsep,topsep=0em]
    \item \textbf{Overall} The optimal decoding method depends on the task, the model, and the priority (e.g., performance vs. robustness vs. speed) in hand. There is no short guideline. The complexity of our results calls for more comprehensive evaluations in future research on decoding methods and careful consideration for practitioners.
    \item \textbf{Performance} The best-performing methods depend on the task at hand. However, some general rules about the divide between different decoding methods still persist in the era of LLMs. Generally, closed-ended tasks favor deterministic methods, while open-ended tasks prefer stochastic methods (\cref{sec:performance}), especially with unaligned models. The performance gap between different decoding methods can be narrowed with alignment. We also provide explanations to understand these phenomena. Moreover, it is also observed that stochastic methods with self-consistency can surpass deterministic ones, albeit requiring multiple runs (\cref{sec:consistency}). 
    \item \textbf{Robustness} The optimal hyperparameters for each decoding method vary according to the model, task, and quantization setting. Some methods achieve superior performance at the cost of exhaustive dataset-specific hyperparameter searches but fail to maintain the superiority when the hyperparameter is fixed. This highlights the \textit{performance-sensitivity} trade-off because LLMs are often confronted with diverse user prompts (\cref{sec:robustness}).
    \item \textbf{Speed} Stochastic decoding and the recently proposed deterministic method, frustratingly simple decoding (FSD)~\cite{yang2023frustratingly}, can achieve a similar decoding speed to greedy search. In contrast, beam search, diverse beam search and other advanced deterministic methods show markedly slower speeds relative to greedy search, with the discrepancy in speed becoming more conspicuous as the length of generation increases for some of those methods  (\cref{sec:speed}).
\end{itemize}

\section{Decoding Methods}
Modern LLMs typically generate text in a left-to-right, token-by-token fashion. For each prefix, the model computes a probability distribution of the next token over a fixed vocabulary. A decoding method defines how the generated token sequence is derived from these probability estimations. We consider decoding methods ranging from deterministic to stochastic. Each method is briefly reviewed below, with detailed descriptions in Appendix~\ref{appx:ds}. The hyperparameter search range of each method is guided by recommendations from relevant literature and common practices.
\subsection{Deterministic Methods}
\paragraph{Greedy Search} selects the token with the highest probability at each time step.
\vspace{-3mm}
\paragraph{Beam Search (BS)}~\cite{freitag-al-onaizan-2017-beam} maintains a beam of the $k$ most probable sequences at each time step, where the hyperparameter $k$ is referred to as the beam width. We consider beam sizes 4 and 8 in our experiments.
\vspace{-3mm}
\paragraph{Diverse Beam Search (DBS)}\cite{vijayakumar2018diverse} is a variant of beam search that divides the $k$ most probable sequences into $G$ groups and incorporates a diversity term to maximize inter-group diversity. In our experiments, we configure various $(k,G)$ pairs of (4,2), (4,4), (8,2), (8,4).
\vspace{-3mm}
\paragraph{Contrastive Search (CS)}~\cite{su2022a} uses a look-ahead mechanism and penalizes tokens compromising the isotropy of the LM's latent space. We search the penalty degree from $\left[0.1,0.2,0.3,0.4,0.5,0.6\right]$ in our experiments.
\vspace{-3mm}
\paragraph{Contrastive Decoding (CD)}~\cite{li2022contrastive} searches for tokens that maximize the probability difference between the LLM and a weaker amateur model. We search the the strength of the amateur penalty from $\left[0.1,0.3,0.5,0.7,0.9\right]$.
\vspace{-3mm}
\paragraph{Frustratingly Simple Decoding (FSD)}~\cite{yang2023frustratingly} exploits the contrasts between the LLM and an auxiliary anti-LM constructed based on the current prefix. There are two variants of FSD: FSD and FSD-d depending on whether the anti-LM is implemented as a vectorized or discrete $n$-gram model. We search penalty degree from  $\left[0.1,0.2,0.3,0.4,0.5,0.6\right]$.
\vspace{-3mm}
\paragraph{DoLa}~\cite{chuang2023dola} obtains the next-token distribution by contrasting the logits differences between the last layer and a premature layer. The premature layer is dynamically selected from a pre-specified set of layers. Following~\citet{chuang2023dola}, we test two sets of layers: even-numbered layers from $\left[0, 16\right)$ and from $\left[16, 32\right)$ respectively.
\subsection{Stochastic Methods}
\label{sec:StochasticMethods}
\paragraph{Temperature Sampling} samples tokens from the estimated next-token distributions. The skewness of distributions can be controlled using a temperature hyperparameter $\tau$. 
We conduct our experiments for $\tau$ within the range of 0.1 to 0.9, incrementing in value of 0.1.
\vspace{-3mm}
\paragraph{Top-$p$ Sampling}~\cite{Holtzman2020The} only considers the minimal set of most probable tokens that cover a specified percentage $p$ of the distribution. We examine across various $p$ thresholds, specifically $\left[0.8,0.85,0.9,0.95,1\right]$.
\vspace{-3mm}
\paragraph{Top-$k$ Sampling}~\cite{fan-etal-2018-hierarchical} only samples from the top-$k$ probable tokens. We explore a range of $k$ values, specifically $\left[5,10,20,50,100\right]$.
\vspace{-3mm}
\paragraph{$\eta$-Sampling}\cite{hewitt2022truncation} truncates words whose probabilities are below an entropy-dependent threshold. The hyperparameter $\eta$ is searched from [3e-4,6e-4,9e-4,2e-3,4e-3].
\vspace{-3mm}
\paragraph{Mirostat Sampling}\cite{basu2021mirostat} directly controls the perplexity rate of the generated text during sampling from top-$k$ tokens ($k$ is determined automatically). We test across a range of log of perplexity values $\tau$ within $\left[2.5,3,4,5\right]$.
\vspace{-3mm}
\paragraph{Typical Sampling}~\cite{meister-etal-2023-locally} sorts the vocabulary according to the differences between the distribution entropy and the token probabilities. In our experiments, we vary the coverage threshold $p$ across the values $\left[0.2,0.9,0.92,0.95\right]$.
\section{Evaluation Setup}
\subsection{Datasets}
Our evaluation spans a variety of tasks.
\vspace{-3mm}
\paragraph{Coding} is an important application of LLMs, facilitating the integration with external tools. We use HumanEval~\cite{chen2021humaneval} and MBPP~\cite{austin2021mbpp}, reporting pass@1 accuracy.
\vspace{-3mm}
\paragraph{Math Problem Solving} is critical for LLMs, enabling them to aid users in numerical reasoning tasks. We employ GSM8K~\cite{cobbe2021gsm8k} for this purpose and report accuracy.
\vspace{-3mm}
\paragraph{Summarization} assists users in capturing the essence of a text. We use CNN/DailyMail (CNN/DM)~\cite{hermann2015CNNDailyMail} and XSUM~\cite{narayan2018xsum}, measuring performance with RougeL~\cite{lin2004rouge}.
\vspace{-3mm}
\paragraph{Translation} is a crucial NLP task to overcome linguistic barriers, thereby facilitating global communication. We benchmark it using four directions of WMT22~\cite{kocmi2022wmt22} and assess the translation quality via BLEU~\cite{papineni2002BLEU}.
\\
\textbf{Commonsense Reasoning} is a key perspective of LLMs for addressing real-world problems. We assess this using CommonsenseQA (CQA)~\cite{talmor2019Commonsenseqa} and StrategyQA (SQA)~\cite{geva2021StrategyQA}, reporting accuracy.
\vspace{-3mm}
\paragraph{Factual Knowledge} is crucial for fulfilling users' informational needs. We measure this using FActScore~\cite{min2023factscore}, reporting on the proportion of correctly generated atomic facts.
\vspace{-3mm}
\paragraph{Instruction Following} reflects the proficiency in responding to diverse user instructions. We use AlpaceEval~\cite{li2023alpacaeval} to compare model performances, using pairwise Win Rate against the reference model, Text-Davinci-003.
\vspace{-3mm}
\paragraph{Open-ended Text Generation} measures the model's capability to produce fluent and coherent content. We utilize datasets including Book~\cite{Zhubook15}, Wikinews\footnote{\url{http://www.wikinews.org}}, and Wikitext~\cite{MerityWikitext17}, and evaluate using MAUVE~\cite{PillutlaMAUVE21}. Notably, open-ended text generation is the primary focus for many recent decoding methods.

For detailed task descriptions and prompts, see Appendix~\ref{appx:ben} and Appendix~\ref{sec:instruction_template}. Generally, higher scores in respective metrics indicate better performance.

\subsection{Models}
We primarily experiment with the Llama-2 family, comprising Llama2 and Llama2-chat~\cite{touvron2023llama2}, representing unaligned and aligned models, respectively. Additional tests include other popular LLMs: MPT~\cite{2023mpt}, CodeLlama~\cite{Rozi2023CodeLlama}, Qwen ~\cite{qwen}, Mistral~\cite{jiang2023mistral}, DeepseekMoE~\cite{dai2024deepseekmoe} and Llama3~\cite{llama3modelcard}, along with their aligned counterparts are detailed in Appendix~\ref{sec:foundation}. Unaligned models are not tested on AlpaceEval and FActScore due to their limited instruction-following capabilities. Owing to poor performance for WMT22 with Llama2-Chat, its performance is measured only on the unaligned model. Unless otherwise specified, we employ half-precision (FP16) for model inference.

\section{Experimental Results}
We perform a thorough evaluation of various decoding methods, assessing them from three critical dimensions. Initially, our analysis centers on the efficacy of these methods across a diverse range of tasks and models. Then, we delve into hyperparameter sensitivity and decoding efficiency.

\begin{table*}[t]
    \vspace{-1em}
    \setlength{\tabcolsep}{3.5pt}
    \small
    \centering
    \resizebox{\textwidth}{!}{\begin{tabular}{c|c|c|cccccccc|cccccc}
    \toprule
    \addlinespace[-0.0001ex]
    \multirow{2}{*}{\textbf{Model}} & \multirow{2}{*}{\textbf{Dataset}}&\multirow{2}{*}{\textbf{Metric}}& \multicolumn{8}{c|}{\textbf{Deterministic Methods}}& \multicolumn{6}{c}{\textbf{Stochastic Methods}} \\
    \cline{4-17}
    \addlinespace[0.2ex]
    &&&\textbf{Greedy} & \textbf{BS} & \textbf{DBS} &\textbf{CS} & \textbf{FSD} &\textbf{FSD-d} &\textbf{CD} &\textbf{DoLa} &\textbf{Temp} &\textbf{Top-$p$} &\textbf{Top-$k$} &\textbf{$\eta$} &\textbf{Miro} & \textbf{Typical} \\
    \cline{1-17}
    \addlinespace[0.5ex] 
    \cline{2-17}
    \addlinespace[0.2ex]
    \multirow{14}{*}[-0.6cm]{\rotatebox{90}{\large Llama2-7B}}
    &HumanEval&\multirow{2}{*}{Pass@1}& 12.80  &\redone 15.24  &\redone 15.24  &14.63  &\redone 15.24  &\redone 15.24  &14.02  &\redone 15.24  &\redone 15.24  &\greenthree 9.15  &\greentwo 8.54  &\greenthree 9.15  &\greenone 7.93  &9.76 \\
    &MBPP&& 17.80  &\redtwo 19.40  &18.40  &17.40  &\redthree 19.20  &\redone 21.20  &18.20  &18.40  &17.20  &14.80  &\greenthree 10.20  &\greentwo 9.40  &\greenone 7.80  &12.00 \\
    \cline{2-17}
    \addlinespace[0.2ex] 
    &GSM8K&Acc& 13.87  &\redtwo 17.21  &\redone 17.74  &14.63  &16.83  &16.60  &\redtwo 17.21  &15.39  &16.30  &12.96  &\greenthree 9.10  &\greentwo 8.64  &\greenone 7.96  &13.04 \\
    \cline{2-17}
    \addlinespace[0.2ex]
    &XSUM&\multirow{2}{*}{R-L}& 27.21  &21.88  &24.65  &\redthree 27.53  &\redtwo 27.75  &\redone 27.88  &27.36  &25.92  &27.14  &22.34  &22.10  &\greentwo 20.45  &\greenone 20.23  &\greenthree 21.33 \\
    &CNN/DM&& \redthree 23.43  &20.69  &21.64  &23.25  &23.39  &\redone 24.05  &\redtwo 23.73  &22.64  &23.40  &20.52  &20.90  &\greentwo 18.63  &\greenone 18.02  &\greenthree 19.13 \\
    \cline{2-17}
    \addlinespace[0.2ex]
    &De$\Rightarrow$En&\multirow{4}{*}{B-4}& \redthree 28.80  &\redone 30.14  &28.71  &28.63  &28.52  &\redtwo 28.82  &28.40  &25.45  &28.55  &22.72  &20.30  &\greentwo 18.44  &\greenone 18.00  &\greenthree 20.00 \\
    &En$\Rightarrow$De&& 22.63  &\redone 23.99  &\redtwo 23.52  &\redthree 22.74  &22.54  &22.63  &22.30  &19.82  &22.57  &16.14  &14.32  &\greentwo 12.28  &\greenone 11.62  &\greenthree 13.34 \\
    &Zh$\Rightarrow$En&& 19.44  &\redone 20.11  &18.90  &19.56  &\redthree 19.71  &\redtwo 20.05  &19.68  &17.06  &19.26  &13.35  &12.02  &\greentwo 10.26  &\greenone 9.60  &\greenthree 10.78 \\
    &En$\Rightarrow$Zh&& 15.15  &14.50  &14.67  &\redtwo 15.27  &\redthree 15.21  &\redone 15.37  &14.57  &13.09  &\redthree 15.21  &11.61  &\greenthree 11.27  &11.50  &\greenone 7.89  &\greentwo 9.94 \\
    \cline{2-17}
    \addlinespace[0.2ex]
    &CQA&\multirow{2}{*}{Acc}& 62.90  &\redone 64.37  &\redtwo 64.21  &63.72  &\redthree 64.05  &63.72  &62.65  &62.00  &63.72  &56.51  &\greenthree 49.47  &\greentwo 47.17  &\greenone 46.11  &52.91 \\
    &SQA&& 60.76  &\redthree 62.25  &61.50  &60.54  &\redtwo 62.90  &60.89  &\redone 63.74  &61.94  &61.20  &58.71  &\greentwo 58.09  &\greenthree 58.27  &58.44  &\greenone 58.05 \\
    \cline{2-17}
    \addlinespace[0.5ex] 
    \cline{2-17}
    \addlinespace[0.2ex]
    &Wikinews&\multirow{3}{*}{MAUVE}& \greentwo 40.10  &\greenthree 41.33  &\greenone 32.02  &96.66  &96.42  &\redtwo 98.40  &85.17  &94.44  &95.40  &95.19  &96.47  &97.48  &\redone 98.51  &\redthree 97.67 \\
    &Wikitext&& \greentwo 23.47  &\greenthree 27.41  &\greenone 22.78  &93.38  &92.14  &92.93  &85.86  &85.39  &\redthree 94.54  &\redtwo 96.62  &\redone 96.67  &93.66  &93.18  &93.29 \\
    &Book&& \greentwo 13.10  &\greenthree 17.54  &\greenone 10.18  &88.41  &89.07  &86.69  &73.30  &80.54  &90.62  &\redone 95.99  &\redthree 94.84  &\redtwo 95.31  &94.25  &93.98 \\
    \cline{2-17}
    \addlinespace[0.5ex]
    \cline{1-17}
    \addlinespace[0.5ex]
    \cline{2-17}
    \addlinespace[0.2ex] 
    \multirow{12}{*}[-0.3cm]{\rotatebox{90}{\large Llama2-7B-Chat}}
    & HumanEval & \multirow{2}{*}{Pass@1}&
    \greentwo 12.80  &14.02  &13.41  &13.41  &\redtwo 15.24  &13.41  &14.02  &\redone 15.85  &\redthree 14.63  &13.41  &14.02  &\greenone 12.20  &\greentwo 12.80  &\greentwo 12.80 \\
    &MBPP &  & 17.20  &\redone 21.60  &\redtwo 21.20  &17.40  &17.80  &17.80  &17.40  &18.00  &\redthree 20.00  &17.60  &\greenone 16.00  &\greentwo 17.00  &\greenone 16.00  &18.00 \\
    \cline{2-17}
    \addlinespace[0.2ex]
    & GSM8K & Acc & 
    24.79  &\redone 28.81  &\redtwo 26.91  &25.70  &25.40  &24.56  &\redthree 26.46  &\greenone 22.14  &25.47  &24.26  &24.41  &25.25  &\greentwo 23.20  &\greenthree 24.11 \\
    \cline{2-17}
    \addlinespace[0.2ex]
    & XSUM & \multirow{2}{*}{R-L} & 16.42  &\redone 16.96  &\redtwo 16.78  &16.70  &16.63  &16.52  &16.49  &\greenone 8.84  &16.51  &16.44  &\greenthree 16.28  &16.44  &\greentwo 15.77  &\redthree 16.77 \\
    & CNN/DM &  & 
    22.59  &\redone 23.71  &\redtwo 23.54  &22.54  &22.40  &22.64  &22.65  &\greenone 16.92  &\redthree 22.71  &22.67  &\greenthree 22.03  &22.34  &\greentwo 20.60  &22.42 \\
    \cline{2-17}
    \addlinespace[0.2ex]
    & CQA & \multirow{2}{*}{Acc} & \greenone 50.61  &\redthree 52.99  &52.83  &\greenthree 51.43  &52.66  &\greentwo 51.11  &52.01  &52.74  &\redone 53.56  &\redtwo 53.15  &51.76  &51.52  &52.66  &52.91 \\
    & SQA &  & 
    59.89  &\redthree 60.41  &\redtwo 60.59  &59.97  &60.32  &60.37  &60.19  &\greenthree 59.62  &60.19  &60.28  &\redone 60.80  &60.10  &\greentwo 59.41  &\greenone 59.14 \\
    \cline{2-17}
    \addlinespace[0.5ex] 
    \cline{2-17}
    \addlinespace[0.2ex]
    & Wikinews & \multirow{3}{*}{MAUVE}&
    \greenone 58.34  &71.01  &74.13  &70.42  &76.74  &\redtwo 81.84  &74.33  &\greentwo 63.99  &\redone 83.84  &76.76  &\redthree 79.65  &72.24  &\greenthree 70.02  &72.32 \\
    & Wikitext & &
    \greentwo 77.69  &87.20  &\redthree 90.27  &\greenthree 80.16  &\redone 95.10  &\redtwo 90.47  &84.59  &\greenone 38.76  &80.45  &80.51  &85.63  &87.52  &83.48  &89.77 \\
    & Book & &
    \greentwo 80.65  &\redthree 94.89  &93.78  &90.81  &94.75  &92.00  &\redtwo 95.96  &\greenone 57.70  &\redone 96.55  &91.50  &93.48  &\greenthree 89.95  &92.95  &93.87 \\
    \cline{2-17}
    \addlinespace[0.5ex] 
    \cline{2-17}
    \addlinespace[0.2ex]
    & FActScore & Score & 44.74  &\redone 47.80  &\redtwo 47.29  &46.09  &46.09  &\redthree 46.93  & 46.11  &\greenone 36.37  &45.06  &44.78  &\greenthree44.11  &46.81  &\greentwo 44.06  &46.55 \\
    \cline{2-17}
    \addlinespace[0.2ex]
    & AlpacaEval & WinRate & \greenthree 76.40& 77.89 & 78.63 &\redthree 79.88& \redtwo 80.50&\redthree 79.88& \redone 81.24  &\greenone 55.40 & 77.76  & 78.01  & 77.39 & 79.38 & \greentwo 75.53 & 78.26  \\
    \bottomrule
    \addlinespace[-0.5ex]
    \end{tabular}}
    \vspace{-1em}
    \caption{Results on Llama2-7B and Llama2-7B-Chat. Cells are colored by performance, from \colorbox{blue_}{low} to medium to \colorbox{orange}{high} performance. The corresponding hyperparameters for each decoding method are listed in Appendix~\ref{appx:hyperparameter}.}
    \label{tab:llama-7b}
\end{table*}
\begin{figure}[t]
\setlength{\abovecaptionskip}{1pt}
    \centering
    \includegraphics[width=0.99\linewidth]{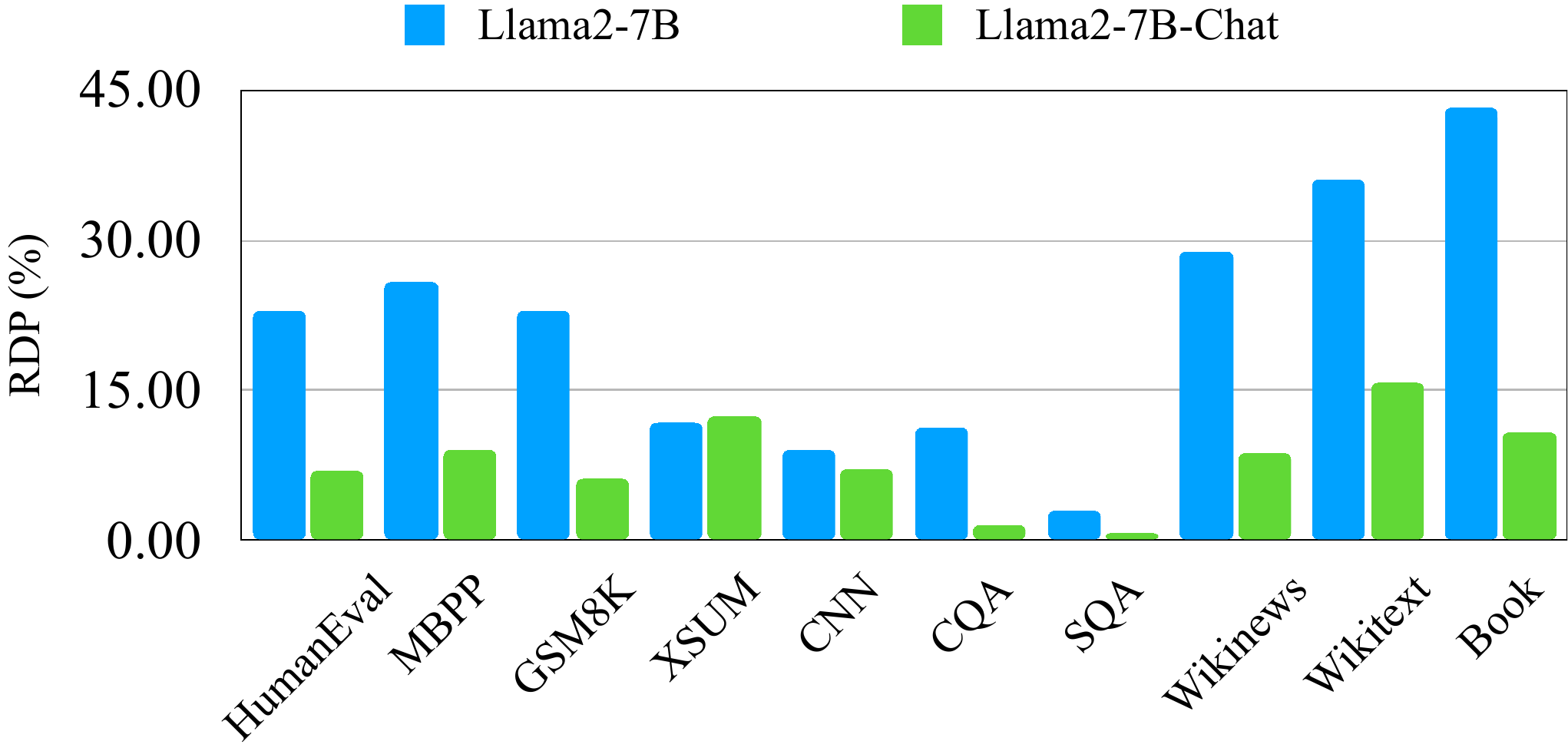}
    \caption{Relative deviation percentage (RDP) for each task on Llama2-7B and Llama2-7B-Chat.}
    \label{fig:std}
    \vspace{-1.5em}
\end{figure}

\subsection{Performance Analysis}
\label{sec:performance}
We present the performance of decoding methods on unaligned and aligned Llama2-7B models (Llama2-7B and Llama2-7B-Chat respectively) in Table~\ref{tab:llama-7b}. The reported results for each method are obtained by utilizing the best hyperparameters tuned for each specific dataset.
\paragraph{For unaligned models, deterministic methods generally perform better than stochastic methods on all tasks except open-ended text generation.}
As shown in the upper block of Table~\ref{tab:llama-7b}, for the unaligned Llama2-7B model, the top-performing decoding methods on closed-ended tasks (coding, math problem solving, summarization, translation, and commonsense reasoning) are frequently among deterministic methods. On the other hand, stochastic methods often struggle with the worst performance. Specifically, BS, FSD-d, and FSD rank in the top 3 (indicated in orange) in 8, 7, and 7 out of 11 datasets, respectively. Conversely, mirostat, $\eta$, and typical sampling are among the least effective three methods (highlighted in blue) in 10, 10, and 7 datasets, respectively. For open-ended text generation (Wikinews, Wikitext, and Book), greedy, BS, and DBS exhibit notably lower MAUVE scores than other methods. The above observations on the disparity of deterministic and stochastic methods are consistent with the findings for conventional task-specific models \citep{wiher2022decoding}: stochastic methods are favorable in open-ended tasks, while heavily disfavored in others.

\noindent$\rightarrow$\textbf{\textit{Phenomenon Analysis.}} Through a careful case study, we find that the outputs of greedy, BS, and DBS contain a considerable amount of repetitive content on open-ended text generation tasks. This suggests that the advanced unaligned LLMs still suffer from the degeneration issue \cite{Holtzman2020The,li2023repetition}. Recent deterministic methods (CS, FSD, FSD-d, CD, and DoLa), which are designed to alleviate the degeneration issue, achieve much better results, performing only slightly worse than stochastic methods. For closed-ended tasks, deterministic approaches are better suited to producing consistent and accurate results as diversity is not a primary concern.


\paragraph{Aligned models are less dependent on decoding methods than unaligned models.}
For the unaligned Llama2-7B model, there is a clear separation between the highest- and lowest-performing methods. For instance, on MBPP, the highest performance is at 21.20\% by FSD-d, in stark contrast to the lowest at 7.80\% by mirostat sampling. However, this distinction becomes less pronounced for the aligned Llama2-7B-Chat model. Specifically, on MBPP, the top performance peaks at 21.60\% while the lowest is at 16.00\%, showcasing a narrowed performance range. 

To further substantiate this, we compute the average $\mu$ and standard deviation $\sigma$ of each dataset across different decoding methods. We report the \textit{relative deviation percentage} (RDP) $\frac{\sigma}{\mu} \times 100\%$, of which a lower value signifies less performance variation across different decoding method choices. The results are depicted in Figure~\ref{fig:std}. Generally, the aligned model (Llama2-7B-Chat) displays less pronounced variations compared to its unaligned counterpart (Llama2-7B), except in two summarization datasets (XSUM and CNN/DM) where the relative deviation percentages are close. This suggests that the choice of decoding method becomes less critical after the model is aligned. Additionally, we also notice that DoLa performs quite worse than other methods under Llama2-7B-Chat. We check its outputs and observe that DoLa fails to terminate its generation appropriately (see Appendix~\ref{appx:dola}).

\noindent$\rightarrow$\textbf{\textit{Phenomenon Analysis.}} The potential reasons are as follows: i) The improved model confidence. As shown in Table~\ref{tab:confidence}, we report the average next-token prediction entropy of Llama2-Chat-7B and Llama2-7B on GSM8K, MBPP, and Wikinews. It can be seen that the entropy of the aligned model is substantially lower than that of the unaligned one. As the model becomes more confident (concentrating the probabilistic mass on a shortlist of tokens), there is less operating space for decoding methods. ii) The alleviated degeneration issue. We find that the aligned model produces much fewer repetitions even when using deterministic decoding methods such as greedy search. This inherent improvement, possibly due to the high-quality data with reduced repetition employed during the instruction tuning phase \citep{li2023repetition}, makes those decoding methods that aim to mitigate the degeneration issue less useful. iii) The more structured writing style. The aligned model typically produces more well-organized responses (e.g., a list of points with explicit discourse markers). This structural coherence enhances the stability of the model's output and reduces the variations of stochastic decoding methods \citep{lin2023unlocking}.
\begin{table}[t]
\centering
\small
\resizebox{0.45\textwidth}{!}{
\begin{tabular}{c|ccc}
\toprule
Model        & GSM8K & MBPP & Wikinews \\
\midrule
Llama2-7B      & 1.05 & 1.21  & 2.37 \\
Llama2-7B-Chat & 0.27 & 0.39  & 0.52 \\
\bottomrule
\end{tabular}}
\vspace{-0.5em}
\caption{The entropy of Llama2-7B and Llama2-Chat-7B's generation results~(top-$p$ sampling with $p=1.0$) on GSM8K, MBPP and Wikinews.}
\vspace{-1.5em}
\label{tab:confidence}
\end{table}

\paragraph{Deterministic methods tend to generate fewer hallucinations and have better instruction-following abilities.}
The lower block of Table~\ref{tab:llama-7b} also presents the results of the aligned model (Llama2-7B-Chat) on FActScore and AlpacaEval. For FActScore, the top three best-performing methods are all deterministic. For instance, beam search attains 47.80\%, while mirostat and top-$k$ sampling only achieve scores of 44.06\% and 44.11\%, respectively. These results indicate that the choice of decoding method has a considerable impact on the factuality of the generated text. The randomness in the selection process of stochastic methods may contribute to increased hallucinations. For AlpacaEval, the general instruction-following task, deterministic methods such as CS, FSD, and CD can outperform all stochastic methods. This observation challenges the prevailing common practice of employing stochastic methods, particularly temperature and top-$p$ sampling, in LLMs. This suggests that deterministic methods are more reliable for tasks requiring high factual accuracy and precise adherence to instructions, warranting further exploration in future research.
\paragraph{Among stochastic methods, temperature sampling generally performs better, particularly when using unaligned models.}
As evidenced in Table~\ref{tab:llama-7b}, temperature sampling generally outperforms other stochastic methods except for open-ended text generation. Specifically, on Llama2-7B, temperature sampling emerges as the top-performing stochastic method across all 11 closed-ended tasks. Similarly, under Llama2-7B-Chat, it takes the top position in 5 out of 9 closed-ended tasks. We find that the best results often come from a low temperature (e.g., $\tau=0.1,0.2$, see Table~\ref{tab:llama-7b-p} in Appendix \ref{appx:hyperparameter}), which renders temperature sampling more akin to deterministic decoding. It is worth noting that many previous studies ~\cite{fan-etal-2018-hierarchical,Holtzman2020The,meister-etal-2023-locally,hewitt2022truncation} predominantly demonstrate the superiority of their proposed methods in the realm of open-ended text generation. However, our analysis reveals that temperature sampling markedly surpasses these methods in closed-ended generation tasks, thereby underscoring the necessity for more holistic evaluations across diverse tasks.
\subsection{Hyperparameter Sensitivity}
\label{sec:robustness}
The results in Table~\ref{tab:llama-7b} are obtained by searching for the optimal hyperparameter of each decoding method for each dataset. Nevertheless, hyperparameter search is time-consuming and may not be plausible for open-world applications where the target task is not known a priori. Therefore, we further explore a more realistic scenario in which each method uses a fixed hyperparameter across different datasets. To ensure a fair comparison that accounts for various performance ranges across different tasks, we first normalize the performance on each dataset according to $normalize(p) = \frac{p}{p_{\text{best}}}\times 100\%$, where $p_{\text{best}}$ represents the best performance obtained in Table \ref{tab:llama-7b}, then compute the average of normalized performance across all datasets, denoted by $\text{ANP}$. We report the best ANP using task-specific hyperparameters ($\text{ANP}_{\text{best}}$) and a fixed hyperparameter ($\text{ANP}_{\text{fix}}$) for each decoding method respectively. The results on Llama2-7B family are presented in Figure~\ref{fig:hyper}. For Llama2-7B, both FSD and FSD-d rank among the top-3 decoding methods in terms of performance, whether under task-specific hyperparameters ($\text{ANP}_{\text{best}}$) or one fixed hyperparameter ($\text{ANP}_{\text{fix}}$), demonstrating that these methods can have the ideal performance without the need for fine-grained selection of hyperparameters for each dataset.
In contrast, while temperature sampling achieves comparable results in terms of $\text{ANP}_{\text{best}}$, it shows an 11.59\% decrease in $\text{ANP}_{\text{fix}}$ when hyperparameters are fixed, highlighting its sensitivity to hyperparameters. Similarly, for Llama2-7B-Chat, BS and DBS perform well and are not sensitive to hyperparameters, while temperature sampling still exhibits a 3.90\% decrease. Notably, CD is also sensitive to hyperparameters, with a performance decrease of 9.42\% on Llama2-7B and 3.35\% on Llama2-7B-Chat.

\definecolor{solidcolor}{HTML}{1F77B4}
\definecolor{hollowcolor}{HTML}{FF7F0E}

\DeclareRobustCommand{\solidcircle}{%
    \tikz[baseline=-0.7ex]\fill[solidcolor] circle (1.35mm); 
}

\DeclareRobustCommand{\hollowcircle}{%
    \tikz[baseline=-0.7ex, line width=0.8pt]\draw[hollowcolor] circle (1.2mm); 
}

\begin{figure}[t]
\setlength{\abovecaptionskip}{1pt}
    \centering
    \includegraphics[width=0.60\linewidth]{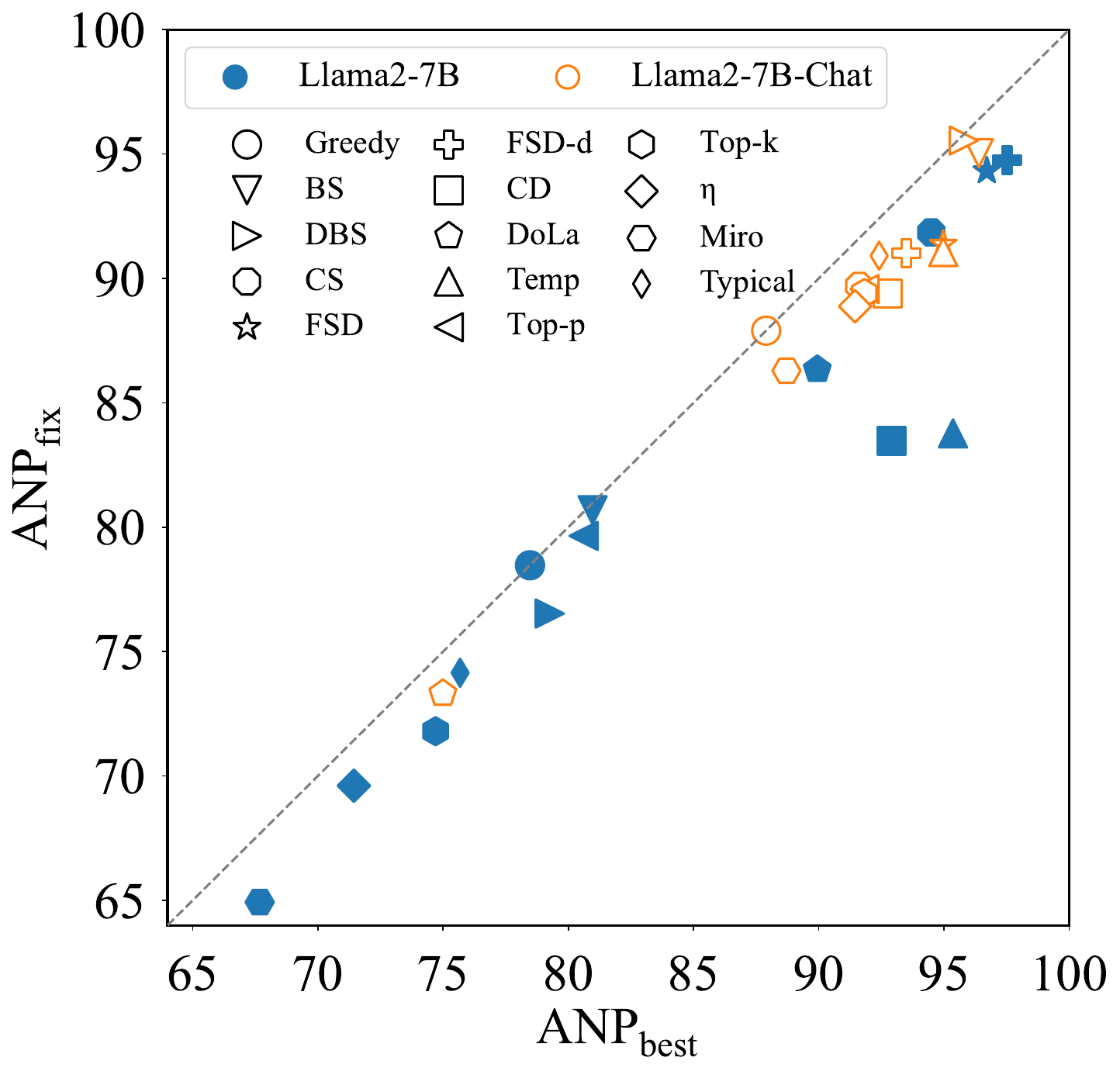}
    \vspace{0.6em}
    \caption{Hyperparameter Sensitivity. $\text{ANP}_{\text{best}}$ and the best $\text{ANP}_{\text{fix}}$ for each decoding method on Llama2-7B with blue solid markers~\solidcircle and Llama2-7B-Chat with orange hollow markers~\hollowcircle{}. The $\text{ANP}_{\text{fix}}$ with the optimal hyperparameters for each decoding method are detailed in Appendix~\ref{appx:hyperparameter}. }
    
    \label{fig:hyper}
    \vspace{-1em}
\end{figure}

\begin{figure}[t]
    \centering
    \includegraphics[width=0.60\linewidth]{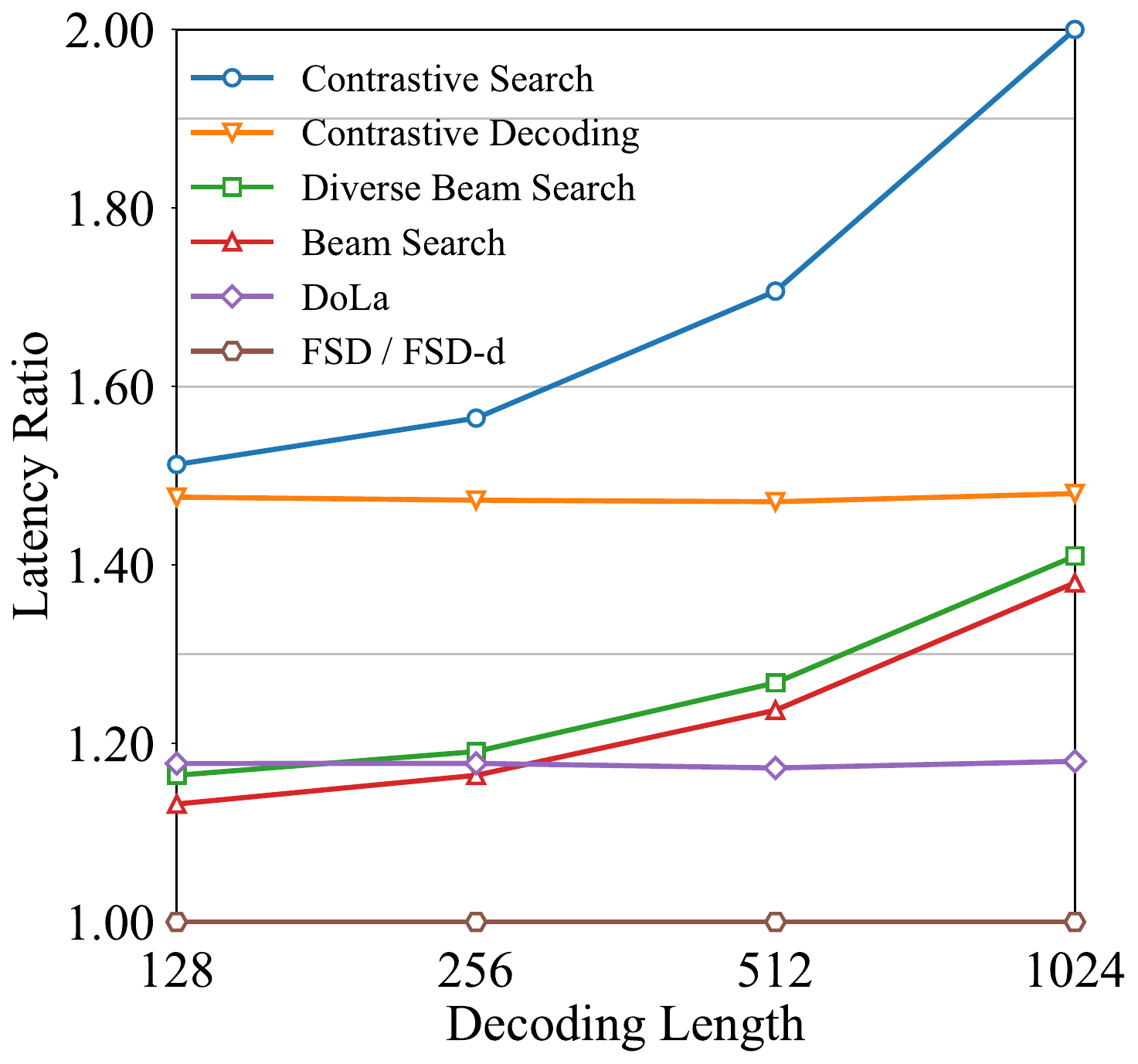}
    \vspace{-0.5em}
    \caption{Decoding latency ratios. The latency is measured on one A6000 GPU with batch size = 1. }
    \label{fig:speed}
    \vspace{-1.5em}
\end{figure}
\subsection{Decoding Speed}
\label{sec:speed}
We assess and compare the decoding speed of various decoding methods in Figure~\ref{fig:speed}. For a more intuitive understanding, we calculate the latency ratio for each decoding method by normalizing their latency with respect to the latency of greedy search. To demonstrate how their latency grows with generation lengths, we plot the latency for generating 128, 256, 512 and 1024 tokens given 32 tokens using Llama2-7B. It is worth noting that we omit the results of all stochastic decoding methods mentioned in \cref{sec:StochasticMethods} because they achieve very close latency to that of greedy search. It is reasonable because their sampling processes only require negligible additional computation.

It can be observed that contrastive search is the decoding method with the slowest decoding speed. Moreover, the latency ratio grows considerably as generation length increases (from 1.51x to 2.00x slower than greedy search). This is due to that the look-ahead mechanism in contrastive search is very time-consuming. Contrastive decoding is about 1.4x slower than greedy search for the additional run of a smaller amateur model. However, the latency ratio of contrastive decoding remains constant across different lengths, indicating better adaptability for long sequence generation. Beam search and diverse beam search are faster than contrastive search and contrastive decoding but slower (1.13x to 1.41x) than greedy search. Both have latency ratios that grow approximately linearly with the sequence length while diverse beam search is slightly slower than beam search. The speed of DoLa is comparable to beam search and diverse beam search when the generation is relatively short (128 and 256). Nevertheless, their difference increases as the generation length grows because the latency ratio of DoLa remains consistent across different lengths. Notably, FSD and FSD-d not only run as fast as greedy search but also maintain a consistent latency ratio across different lengths, underscoring their superior efficiency against other advanced deterministic decoding methods.

\section{Further Analysis}
\begin{figure}[t]
    \centering
    \includegraphics[width=0.99\linewidth]{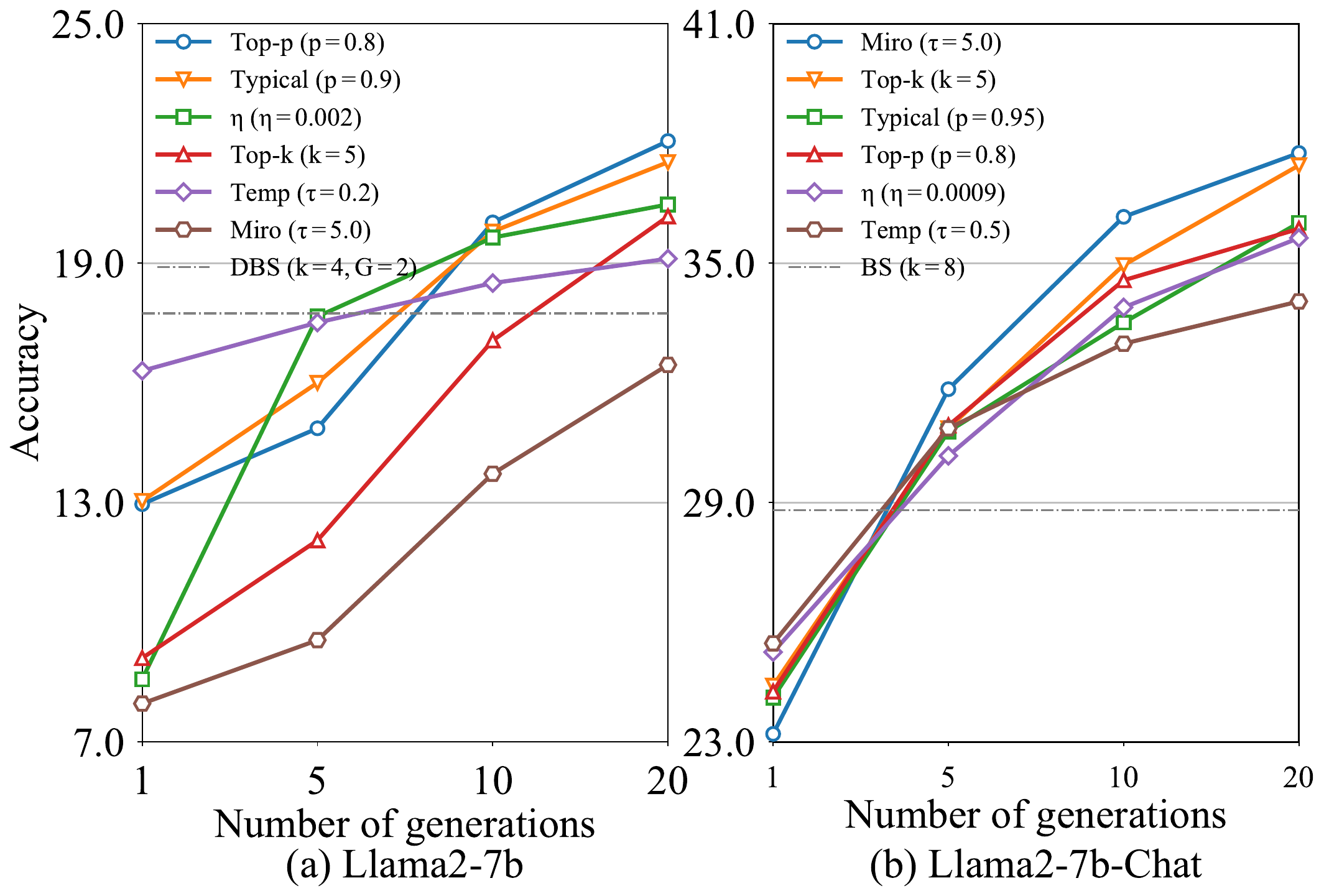}
    \caption{Results of stochastic decoding methods with self-consistency on GSM8K.}
    \label{fig:3}
\end{figure}
\begin{table}[t]
    \centering
    \resizebox{0.45\textwidth}{!}{
    \small
    \begin{tabular}{c|cccccc}
    \toprule
    Model & Temp & Top-$p$ & Top-$k$ & $\eta$ & Miro & Typical \\
    \midrule
    \multirow{2}{*}{7B} & 21.91 & 22.06 & 20.17 & 21.23 & 16.98 & 22.06 \\
    & (0.7) & (0.8) & (5) & (0.004) & (4.0) & (0.95) \\
    \midrule
    \multirow{2}{*}{7B-Chat} & 36.92 & 36.85 & 37.68 & 35.63 & 37.76 & 36.92 \\
     & (0.9) & (1.0) & (10) & (0.0009) & (5.0) & (0.90) \\
    \bottomrule
    \end{tabular}
    }
    \vspace{-0.5em}
    \caption{Best results of different stochastic methods with self-consistency~(20 generations) setting on GSM8K for Llama2-7B family. The best hyperparameters are annotated in parentheses.
    }
    \vspace{-1.5em}
    \label{tab:self}
\end{table}
\begin{table*}[t]
    \small
    \centering
    \resizebox{0.95\textwidth}{!}{\begin{tabular}{c|c|cccccccc|cccccc|c}
    \toprule
    \addlinespace[-0.0001ex]
    \multirow{2}{*}{\textbf{Model}} & \multirow{2}{*}{\textbf{Dataset}}& \multicolumn{8}{c|}{\textbf{Deterministic Methods}}& \multicolumn{6}{c|}{\textbf{Stochastic Methods}} & \multirow{2}{*}{\textbf{RDP}}\\
    \cline{3-16}
    \addlinespace[0.2ex]
    &&\textbf{Greedy} & \textbf{BS} & \textbf{DBS} &\textbf{CS} & \textbf{FSD} &\textbf{FSD-d} &\textbf{CD} &\textbf{DoLa} &\textbf{Temp} &\textbf{Top-$p$} &\textbf{Top-$k$} &\textbf{$\eta$} &\textbf{Miro} & \textbf{Typical}& \\
    \cline{1-17}
    \addlinespace[0.2ex]
    \multirow{3}{*}{7B} &MBPP& 17.80  &\redtwo 19.40  &18.40  &17.40  &\redthree 19.20  &\redone 21.20  &18.20  &18.40  &17.20  &14.80  &\greenthree 10.20  &\greentwo 9.40  &\greenone 7.80  &12.00 &25.81\\
    &GSM8K& 13.87  &\redtwo 17.21  &\redone 17.74  &14.63  &16.83  &16.60  &\redtwo 17.21  &15.39  &16.30  &12.96  &\greenthree 9.10  &\greentwo 8.64  &\greenone 7.96  &13.04 &23.06 \\
    &Wikinews& \greentwo 40.10  &\greenthree 41.33  &\greenone 32.02  &96.66  &96.42  &\redtwo 98.40  &85.17  &94.44  &95.40  &95.19  &96.47  &97.48  &\redone 98.51  &\redthree 97.67 &28.83\\
    \cline{1-17}
    \addlinespace[0.2ex]
    \multirow{3}{*}{13B} &MBPP&23.00  &\redthree 24.00  &23.20  &\redtwo 24.40  &23.00  &\redone 25.80  &23.00  &23.80  &23.40  &17.40  &\greentwo 13.40  &21.60  &\greenone 10.00  &\greenthree 17.20 &21.37\\
    &GSM8K & 
    28.81  &29.64  &29.19  &29.42  &\redtwo 31.99  &\redthree 31.16  &\redone 33.36  &28.58  &30.02  &24.94  &\greentwo 18.20  &30.10  &\greenone 15.39  &\greenthree 21.76 &18.74\\
    &Wikinews & \greenthree 62.02  &\greenone 50.30  &\greentwo 51.00  &\redone 98.22  &97.01  &93.26  &94.83  &91.53  &96.88  &\redthree 97.77  &\redtwo 97.81  &97.19  &96.94  &96.87 &19.95\\
    \cline{1-17}
    \addlinespace[0.2ex]
    \multirow{3}{*}{70B} &MBPP&41.80 & \redone 43.40 & 41.00 & 39.40 & 41.20 & 41.20 & \redtwo 42.20 & 37.00 & 41.80 & \greenthree 33.20 & \greentwo 25.80 & 38.80 & \greenone 24.80 & \redtwo 42.20 &15.23\\
    &GSM8K&57.39 & 59.44 & 58.76 & 58.91 & \redthree 60.73 & 60.42 & \redone 63.91 & \redtwo 61.33 & 57.47 & \greenthree 53.37 & \greentwo 44.20 & 58.53 & \greenone 38.36 & 59.89 &11.92\\
    &Wikinews &\greenone42.44 &\greentwo76.35 &\greenthree77.3 3&95.22 &\redthree 95.68 &93.29 &95.3 &94.31 &94.09 &92.75 &93.39 &\redone96.04 &\redtwo 96.02& 92.33 &16.02\\    
    \cline{1-17}
    \addlinespace[0.2ex]
    \multirow{3}{*}{7B-Chat} &MBPP& 17.20  &\redone 21.60  &\redtwo 21.20  &17.40  &17.80  &17.80  &17.40  &18.00  &\redthree 20.00  &17.60  &\greenone 16.00  &\greentwo 17.00  &\greenone 16.00  &18.00 &9.08\\
    &GSM8K & 24.79  &\redone 28.81  &\redtwo 26.91  &25.70  &25.40  &24.56  &\redthree 26.46  &\greenone 22.14  &25.47  &24.26  &24.41  &25.25  &\greentwo 23.20  &\greenthree 24.11 &6.25\\
    &Wikinews & \greenone 58.34  &71.01  &74.13  &70.42  &76.74  &\redtwo 81.84  &74.33  &\greentwo 63.99  &\redone 83.84  &76.76  &\redthree 79.65  &72.24  &\greenthree 70.02  & 72.32 &8.84\\
    \cline{1-17}
    \addlinespace[0.2ex]
    \multirow{3}{*}{13B-Chat} &MBPP& \greenone 22.60  &\redone 24.80  &\redtwo 24.40  &23.80  &24.00  &\greentwo 23.40  &23.80  &23.60  &\redone 24.80  &24.00  &24.00  &24.20  &\greenone 22.60  &23.60 &2.69\\
    &GSM8K&\greentwo 34.57  &\redone 39.73  &\redtwo 38.06  &36.24  &36.62  &36.16  &36.62  &\greenone 33.13  &36.32  &36.01  &\greenthree 35.41  &\greenthree 35.41  &36.01  &\redthree 36.85 &4.05 \\
    &Wikinews& \greentwo 77.35  &84.43  &88.82  &87.80  &\redtwo 92.89  &82.58  &\redone 98.06  &\greenone 70.68  &84.54  &87.50  & \greenthree 82.20  &89.20  &89.23  &\redthree 90.13 &7.50\\
    \cline{1-17}
    \addlinespace[0.2ex]
    \multirow{3}{*}{70B-Chat} &MBPP& 31.40  &\redtwo 31.80  &\redone 32.00  &30.40  &30.80  &30.80  &30.60  &\greenthree 30.20  &\redone 32.00  &30.80  &\greentwo 28.40  &31.60  &\greenone 28.20  &31.60 &3.74 \\
    &GSM8K&51.93 & \greentwo 50.87 & \redone 53.90 & \redtwo 53.22 & 52.01 & 52.54 & 52.24 & \greenone 48.82 & 52.62 & \redthree 52.99 & \greenthree 51.10 & 52.92 & 51.93 & 52.16 &2.28\\
    &Wikinews& \greenthree77.53& \greentwo74.01&84.10& \redthree85.85& 84.60 &\redthree 87.13 &81.54&\greenone69.58&80.67&82.00&83.85&82.53 &75.11& \redthree 84.69 &6.04\\
    \addlinespace[-0.5ex]
    \bottomrule
    \end{tabular}}
    \vspace{-0.5em}
    \caption{Results of Llama2 family models with different scales on MBPP, GSM8K, Wikinews datasets. We report the relative deviation percentage (RDP) of the performance of different decoding methods on each task in the last column. The corresponding hyperparameters for each decoding method are listed in Appendix~\ref{appx:hyperparameter}.  }
    \vspace{-0.5em}
    \label{tab:scale}
\end{table*}
\subsection{Self-Consistency} 
\label{sec:consistency}
Previous experiments demonstrate that the best-performing decoding methods are generally deterministic ones on closed-ended tasks, particularly on complex reasoning tasks such as the GSM8K dataset. Nonetheless, one unique advantage of stochastic decoding methods is that they can produce varied results through multiple runs, of which one can use the self-consistency strategy~\cite{wang2023selfconsistency} for enhanced task performance. Concretely, self-consistency samples multiple generations and takes a majority vote to determine the final answer. To gain further insights into the potential of stochastic decoding methods, we then delve into the experiments with self-consistency. 

As illustrated in Figure~\ref{fig:3}, we plot the accuracies of various stochastic decoding methods on GSM8K with respect to varying numbers of sampled generations (1, 5, 10, and 20). We also contrast the results with the best accuracies achieved by deterministic decoding methods (i.e., 17.74\% by diverse beam search using Llama2-7B and 28.81\% by beam search using Llama2-7B-Chat), denoted by the gray dashed lines. The results show that sampling a larger number of generations consistently leads to better performance, confirming the usefulness of self-consistency in taking advantage of the diversity introduced by stochastic sampling. Except for the results of mirostat sampling on Llama2-7B, we can see that all stochastic methods eventually surpass the best-performing deterministic methods when the number of sampling reaches 20. Note that the results in Figure~\ref{fig:3} are obtained by using the best hyperparameters we find in Table~\ref{tab:llama-7b} where only one-pass generation is allowed. 

We speculate that further tuning the hyperparameters can improve the performance under the self-consistency strategy. Thus we undertake an additional hyperparameter search in scenarios where the number of generations is set to 20. The highest results along with the corresponding hyperparameters are reported in Table~\ref{tab:self}. Compared to the results in Figure~\ref{fig:3}, we can see that the performance is boosted by employing a hyperparameter with greater randomness or candidate pool. For example, on Llama2-7B-Chat, the accuracy of temperature sampling increases from 34.04\% ($\tau=0.5$) to 36.92\% ($\tau=0.9$). Another interesting finding is that the best hyperparameters for aligned models typically suggest greater randomness (e.g., $\tau=0.9$ vs. $\tau=0.7$ for temperature sampling).
\subsection{Scaling Model Size}
\begin{table*}[th]
    \small
    \centering
    \resizebox{0.99\textwidth}{!}{\begin{tabular}{c|c|cccccccc|cccccc|c}
    \toprule
    \addlinespace[-0.0001ex]
    \multirow{2}{*}{\textbf{Model}} & \multirow{2}{*}{\textbf{Dataset}}& \multicolumn{8}{c|}{\textbf{Deterministic Methods}}& \multicolumn{6}{c|}{\textbf{Stochastic Methods}}& \multirow{2}{*}{\textbf{RDP}}\\
    \cline{3-16}
    \addlinespace[0.2ex]
    &&\textbf{Greedy} & \textbf{BS} & \textbf{DBS} &\textbf{CS} & \textbf{FSD} &\textbf{FSD-d} &\textbf{CD} &\textbf{DoLa} &\textbf{Temp} &\textbf{Top-$p$} &\textbf{Top-$k$} &\textbf{$\eta$} &\textbf{Miro} & \textbf{Typical}& \\
    \cline{1-17}
    \addlinespace[0.2ex]
    \multirow{3}{*}{13B-INT4}&MBPP&  23.00 &\redtwo 24.60 & 23.00 &21.20 &\redtwo 24.60 &\redone 25.20 &23.00 &23.00 & 24.00 & \greenthree 18.40 &\greentwo 11.40 &21.00 & \greenone10.40 &22.60 &21.28\\
    &GSM8K &  27.45 &\redthree 30.33 & 28.13 &27.67 &\redtwo 31.01 & 30.10 &\redone 31.46 & 29.11 &27.60 & \greenthree 21.30 &\greentwo 15.69 &26.38 & \greenone 14.10 &27.90&19.92\\
    &Wikinews & \greenthree47.41 &\greenone46.61 &\greentwo46.70 &91.21 & 96.11 & 95.91& 87.51& 92.83&\redtwo 97.79 &\redone 97.99 &\redthree 96.70 &90.32 & 95.67 &91.02&23.28\\
    \cline{1-17}     \addlinespace[0.2ex]
    \multirow{3}{*}{13B-INT8} &MBPP& 21.60 & \redthree23.20 & 22.20 &\redthree 23.20 & 22.60 & \redone 25.20 & 22.80 & \redtwo 24.00 & 23.00 & 17.60 & \greenthree 12.60 & \greentwo11.80 & \greenone9.60 & 15.40 &25.46\\
    &GSM8K & 28.43 & 28.89 & 28.96 & 29.04 & \redtwo30.93 & \redthree30.48 & \redone33.59 & 28.28 & 29.34 & 23.88 & \greenthree17.21 & \greentwo16.45 & \greenone13.34 & 21.68&23.25 \\ &Wikinews& \greentwo 49.24& \greenthree 51.92& \greenone 45.56&94.39&96.99& \redthree 97.18&93.96&94.06&94.81&\redone97.71&96.33&\redthree97.28&95.60&97.15&22.62\\
    \cline{1-17}     \addlinespace[0.2ex]
    \multirow{3}{*}{13B-Chat-INT4}&MBPP& 23.80 &\redtwo 25.60 &\redone 25.80 &\redthree 25.40 & 24.80 &24.60 &24.40 & \greenthree22.80 & 24.80 &24.40 & \greenone21.60 &24.00 &\greentwo 22.40 &\redthree 25.40&4.97\\
    &GSM8K &  34.12 &\redtwo 35.71 &\redone 37.45 &34.50 & 35.33 &\redthree 35.41 &34.42 & \greenone 31.61 & 35.33 &34.27 &\greenthree33.97 &34.04 & \greentwo33.74 & 35.33&3.64 \\
    &Wikinews &\greentwo80.34 &83.65 &83.16 & 86.81 & 85.34 & 86.98&\redone 91.40& \greenone73.72 &\redthree 87.76 &\redtwo 89.29 &\greenthree81.25 & 84.63 &83.02 & 83.90&4.93\\
    \cline{1-17}     \addlinespace[0.2ex]
    \multirow{3}{*}{13B-Chat-INT8} &MBPP&  24.00 & 23.80 & \redtwo24.60 & \redthree24.20 & \greentwo 22.80 & \greenone22.40 & 23.40 & 23.60 & 23.40 & \greenthree 23.20 & 23.40 & \redone25.20 & 23.40 & 23.80&2.88 \\
    & GSM8K &  \greentwo 35.56 & 36.92 & \redthree 37.68 & \redtwo 37.76 & 36.69 & 36.69 & 37.38 & \greenone 31.54 & \greenthree 36.09 & \redone 38.44 & 37.15 & 36.92 & 36.85 & 37.38 &4.29\\
&Wikinews&\greentwo73.73&83.22&\redthree88.82&\redone91.37&85.53&\greenthree81.25&89.41&\greenone57.87&\redtwo90.50&88.67&87.10&82.91&84.87&81.61&10.34\\
    \addlinespace[-0.5ex]
    \bottomrule
    \end{tabular}}
    \vspace{-0.5em}
    \caption{Results for INT4 and INT8 quantization with Llama2 13B family on MBPP, GSM8K, Wikinews datasets. The corresponding hyperparameters for each decoding method are listed in Appendix~\ref{appx:hyperparameter}.}
    \vspace{-1.5em}
    \label{tab:quantization}
\end{table*}
\label{sec:size}
In order to investigate the impact of model scale on different decoding methods, we provide further experiments on Llama2 family with 13B and 70B parameters in 3 representative tasks: MBPP, GSM8K, and Wikinews. We present the results in Table~\ref{tab:scale}. 
It can be observed that as the model's parameters increase, the \textit{relative deviation percentage} (RDP) of each task decreases, indicating that the differences between different decoding methods have been reduced. 
This suggests that scaling model size can diminish the significance of decoding strategies. 
Moreover, as the number of model parameters varies, the optimal hyperparameters for each decoding method are also subject to change (detailed in Appendix~\ref{appx:hyperparameter}). Consequently, there is also a need to adjust the hyperparameters for larger-scale models individually, rather than directly applying those from smaller models.
Meanwhile, the degree of impact from the model scale varies for different decoding methods. For example, in MBPP, the best performance of $\eta$ sampling on Llama2-7B is 9.40\%, which is less than half of the best method FSD-d at 21.20\%. However, for Llama2-13B, $\eta$ sampling achieves 21.60\%, and for Llama2-70B, it reaches 38.80\%, showing comparable efficacy to the best decoding method. This shows $\eta$ sampling benefits greatly from a greater model scale.
\subsection{Quantization}
\label{sec:quant}
The large size of LLMs presents challenges for deployment, especially where resources are limited. Consequently, in the LLM era, it is crucial to examine how various decoding methods perform in quantization settings. We assess the performance of decoding methods in both INT8 quantization~\cite{dettmers2022llmint8} and INT4 quantization~\cite{lin2023awq} for Llama2-13B family. As detailed in Table~\ref{tab:quantization}, compared with the FP16 13B model in Table~\ref{tab:scale}, the RDP under quantized models is larger, indicating that quantization may impact the models' robustness to different decoding methods. At the same time, different decoding methods exhibit varying adaptability to quantized models. Specially, the performance changes of deterministic methods before and after quantization are not significant for both INT4 and INT8. However, for $\eta$ and typical sampling, there are noticeable changes when quantizing Llama2-13B. Taking GSM8K as an example, $\eta$ sampling under INT8 quantization decreased by 13.65\%, while typical sampling under INT4 quantization improved by 6.14\% on GSM8K. Typical and $\eta$ sampling are more influenced by quantization because their computing involves numerically unstable calculations. This may indicate that the impact of quantization on the information entropy of different tokens during decoding cannot be ignored. 
\section{Conclusion}
This study offered a comprehensive analysis of diverse traditional and contemporary decoding methods in the context of LLMs. Our experiments shed light on the efficacy, robustness, efficiency, and universality of these decoding methods across a range of tasks, models, and settings. One primary finding is that the choice of decoding methods remains crucial and different decoding methods manifests different advantages in different scenarios. However we still provide some practical guidelines in Appendix~\ref{takeaway}. We hope this investigation provides valuable insights and guidance for practitioners and researchers in selecting and advancing decoding methods for LLMs.
\section*{Limitations}
Despite the thoroughness of our study, there are some inherent limitations. First, while we have explored a variety of tasks and models, the ever-evolving nature of LLMs implies that new models or tasks might display distinct behaviors. Second, although our analysis of hyperparameter sensitivity covers a wide range of commonly used configurations, it is not exhaustive and does not account for all possible hyperparameters. Lastly, this paper does not explore the integration of multiple decoding methods, such as combining temperature sampling with a repetition penalty mechanism.
\section*{Acknowledgments}
This research is partly supported by the Shenzhen Science and Technology Program (JCYJ20220818101014030) and the "Graph Neural Network Project" of Ping An Technology (Shenzhen) Co., Ltd. Additionally, the work described in this paper is substantially funded by a grant from the Research Grant Council of the Hong Kong Special Administrative Region, China (Project Code: 14200620).
\bibliography{acl_latex}
\clearpage
\appendix
\section{Decoding Strategies}
\label{appx:ds}
\subsection{Deterministic Methods}
\paragraph{Greedy Search} is arguably the simplest decoding strategy. At each time step $t$, it selects the token with the highest probability predicted by the model from the whole vocabulary set $\mathcal{V}$. Mathematically, the chosen token $y_t$ at time $t$ is:
\begin{equation}
    y_t = \argmax_{y \in \mathcal{V}} P(y|\mathbf{x},\mathbf{y}_{<t})
\end{equation}
where $\mathbf{x}$ is the original input and  $\mathbf{y}_{<t}$ is the generated tokens until time $t-1$. 
One drawback of greedy search is that it does not consider the global sequence score and can get stuck in local optima. This is why beam search is devised.

\paragraph{Beam Search}~\cite{freitag-al-onaizan-2017-beam} maintains a set, or "beam", of the $k$ most probable sequences at each time step, where the hyperparameter $k$ is referred to as the beam width. 
At time $t$, for each $\mathbf{y}_{<t} \in \mathcal{B}_{t-1}$, where $\mathcal{B}_{t-1}$ is the set of $k$ most probable sequences at time $t-1$, it calculates a score for each token $y \in \mathcal{V}$: 
\begin{equation}
    \score(\mathbf{y}_{<t}, y) = \log P(\mathbf{y}_{<t}, y| \mathbf{x}) 
\end{equation}
Then, a new set $\mathcal{B}_t$ is obtained:
\begin{equation}
    \mathcal{B}_t = \argtopk_{\mathbf{y}_{<t} \in \mathcal{B}_{t-1}, y\in \mathcal{V}} {\score(\mathbf{y}_{<t}, y)}
\end{equation}
We specifically test beam size 4 and 8 in our experiment.

\paragraph{Diverse Beam Search}~\cite{vijayakumar2018diverse} is a variant of beam search and aims to improve the diversity among the generated sequences. It divides the $k$ sequences into $G$ groups, each with a size of $k/G$ sequences. The algorithm operates in a similar way to the standard beam search, but instead of choosing the top-$k$ sequences from all candidate sequences, it selects the top- $k/G$ sequences for each group. The key difference lies in how the scores are calculated. In diverse beam search, a penalty is added to the score of a sequence if a similar sequence has already been in other groups:

\begin{equation}
\begin{split}
    \score(\mathbf{y}_{<t}, y)_{\mathbf{y}_{<t} \in \mathcal{B}_{t-1}^g} = \log P(\mathbf{y}_{<t}, y| \mathbf{x}) \\ - \lambda \sum_{g'<g} \Delta ((\mathbf{y}_{<t}, y), \mathcal{B}_{t}^{g'})
\end{split}
\end{equation}
where $\Delta ((\mathbf{y}_{<t}, y), \mathcal{B}_{t}^{g'})$ is a measure of similarity between $(\mathbf{y}_{<t}, y)$  and sequences within $\mathcal{B}_{t}^{g'}$. In our experimental setup, we configure various $(k,G)$ pairs of (4,2), (4,4), (8,2), (8,4), and the diversity penalty $\lambda$ is always set to 1. 

\paragraph{Contrastive Search} ~\cite{su2022a} assumes the LM has an isotropic representation space and adds a penalty term that decreases the generation probabilities of tokens producing hidden states that are very similar to the previous context. Formally, given the context $(\mathbf{x},\mathbf{y}_{<t})$, the selection of the output $y_t$ follows
\begin{equation}
\begin{split}
    y_t = \argmax_{y\in V^k}(1-\alpha)P(y|\mathbf{x},\mathbf{y}_{<t})\\ - \alpha \max\left\{s(h_y,h_v): v\in (\mathbf{x},\mathbf{y}_{<t})\right\}
\end{split}
\end{equation}
where $\mathcal{V}^k$ is the set of top-$k$ predictions from the language model’s probability distribution $P(y|\mathbf{x},\mathbf{y}_{<t})$.
$h_v$ is the hidden states for the token $v$, and $s$ is the similarity function where the cosine similarity is usually adopted. We search $\alpha$ from $\left[0.1,0.2,0.3,0.4,0.5,0.6\right]$ in our experiment.

 \paragraph{Contrastive Decoding} ~\cite{li2022contrastive} employs an additional amateur LM and penalizes undesired attributes associated with the amateur model. Formally, for each candidate token $y \in \mathcal{V}^c$
\begin{equation}
     \score((\mathbf{x},\mathbf{y}_{<t}),y)= (1+\beta)* \mathbf{u}_y - \beta *\mathbf{v}_y
 \end{equation}
$\mathbf{u}$ and $\mathbf{v}$ are the logits before softmax of the expert and amateur models respectively. These two models have the same tokenizer and the expert model is usually much larger than the amateur model.
$\mathcal{V}^c$ is a set of candidate tokens selected based on the following criteria:
\begin{equation}
\begin{split}
    &\mathcal{V}^c(\mathbf{x},\mathbf{y}_{<t}) =  \\ &\left\{ y\in \mathcal{V}: P_{exp}(y|\mathbf{x},\mathbf{y}_{<t}) > \alpha \max P_{exp}(\cdot|\mathbf{x},\mathbf{y}_{<t})\right\}
\end{split}
\end{equation}
In our experiment, we adopt TinyLlama-1.1B\footnote{\url{https://huggingface.co/TinyLlama/TinyLlama-1.1B-intermediate-step-955k-token-2T}} as the amateur model. We use the default setting with $\alpha$ set to 0.1 and we search $\beta$ from $\left[0.1,0.3,0.5,0.7,0.9\right]$.

\paragraph{Frustratingly Simple Decoding}~\cite{yang2023frustratingly} exploits the contrasts between the LLM and an auxiliary anti-LM constructed based on the current prefix. There are two variants of FSD: FSD and FSD-d depending on whether the anti-LM is implemented as a vectorized or discrete $n$-gram model. Specifically, the FSD score is defined as

 \begin{equation}
 \begin{split}
     \mathrm{FSD}(y|\mathbf{x}, \mathbf{y}_{< t}) = (1-\alpha)P_{\theta}(y|\mathbf{x}, \mathbf{y}_{< t})- \\ \alpha \times P_{\omega}(y|\mathbf{x}, \mathbf{y}_{< t})
\end{split}
 \end{equation}
where $P_\theta$ and $P_\omega$ represent the LM and the anti-LM respectively. The hyper-parameter $\alpha \geq 0$ is used to balance the two scores. In practice, it first selects the top-$k$ most probable tokens according to $P_\theta (\cdot|\mathbf{x}, \mathbf{y}_{< t})$, denoted by $\mathcal{V}^{k}$. The token in $\mathcal{V}^{(k)}$ with the largest $\mathrm{FSD}$ score is chosen as the $t_{th}$ token. We search $\alpha$ from  $\left[0.1,0.2,0.3,0.4,0.5,0.6\right]$.

\paragraph{DoLa}~\cite{chuang2023dola} obtains the next-token distribution by contrasting the logits differences between the last layer and a premature layer. For Llama2-7b, the premature layer is dynamically selected from even-numbered layers from $\left[0, 16\right)$ and $\left[16, 32\right)$. For Llama2-13b, the ranges are $\left[0, 20\right)$ and $\left[20, 40\right)$. For Llama2-70b, the ranges are $\left[0, 20\right)$ and $\left[60, 80\right)$. They adopt the Jensen-Shannon divergence (JSD) as the measure of distance between the next-word distributions and select the layer that has the largest JSD as the premature layer. 

\subsection{Stochastic Methods}
\paragraph{Temperature Sampling}  is a decoding strategy to control the randomness in the sampling process. Instead of directly sampling tokens from the predicted distribution, temperature sampling introduces a hyperparameter "temperature" $\tau$ that is used to adjust the probability distribution:
\begin{equation}
    P(y |\mathbf{x}, \mathbf{y}_{<t}) = \frac{\exp(\mathbf{u}_y/\tau)}{\sum_j \exp(\mathbf{u}_j/\tau)}
\end{equation}
where $\mathbf{u}_y$ is the logit of $y$ before softmax. We conduct our experiment for $\tau$ within the range of 0.1 to 0.9, incrementing in value of 0.1.

\paragraph{Top-$k$ Sampling}~\cite{fan-etal-2018-hierarchical}  is used to ensure that the less probable words, which are in the unreliable tail of the distribution~\cite{Holtzman2020The}, should not have any chance to be selected. Only top-$k$ probable tokens are considered for a generation.  
we explore a range of $k$ values, specifically $\left[5,10,20,50,100\right]$.

\paragraph{Top-$p$ Sampling}~\cite{Holtzman2020The} considers the minimal set of top tokens $\mathcal{V}^p$ that cover a specified percentage $p$ of the distribution:
\begin{equation}
    \sum_{y\in \mathcal{V}^p}P(y|\mathbf{x}, \mathbf{y}_{<t})\geq p 
\end{equation}
For our study, we have examined various $p$ thresholds, specifically $\left[0.8,0.85,0.9,0.95,1\right]$.

\paragraph{Typical Sampling} ~\cite{meister-etal-2023-locally} sorts the vocabulary according to the differences between distribution entropy and probabilities. The authors argue that the desired sequences should have information content close to the expected information content, i.e., the conditional entropy of the model. The candidate set $\mathcal{V}^c$ is a solution of the following problem:
\begin{equation}
\begin{split}
    \min_{\mathcal{V}^c}\sum_{y \in \mathcal{V}^c}|H(Y_t|&\mathbf{x},\mathbf{y}_{<t})+\log P(y|\mathbf{x},\mathbf{y}_{<t})| \\
    &\text{s.t.} \sum_{y \in \mathcal{V}^c}P(y|\mathbf{x},\mathbf{y}_{<t})\geq p 
\end{split}
\end{equation}
In our experiments, we vary the threshold $p$ across the values $\left[0.2,0.9,0.92,0.95\right]$ to examine its effect on sequence generation.

\paragraph{Top-$\eta$ Sampling}\cite{hewitt2022truncation} truncates words whose probabilities are below an entropy-dependent threshold. The candidate set $\mathcal{V}^c$ is determined by:
\begin{equation}
\begin{split}
    &\mathcal{V}^c = \\&\left\{ y\in \mathcal{V}|P (y|\mathbf{x}, \mathbf{y}_{<t})\geq \sqrt{\eta} \exp(-h_{\theta,(\mathbf{x},\mathbf{y}_{<t})})\right\} 
    \\
\end{split}
\end{equation}
where $h_{\theta,(\mathbf{x},\mathbf{y}_{<t})}$ is the entropy of $P(Y|\mathbf{x},\mathbf{y}_{<t})$.
$\eta$ is searched from $\left[0.0003,0.0006,0.0009,0.002,0.004\right]$.
\paragraph{Mirostat Sampling}\cite{basu2021mirostat} directly control the perplexity rate of the generated text. It firstly estimates the value
of $s$ assuming words follow Zipf’s law where $s$ is an exponent characterizing the distribution. Then it uses top-$k$ sampling to generate the new token where $k$ is a function of the estimated $s$ and of the target perplexity $\tau$ of the output text. We search $\tau$ from $\left[2.5,3,4,5\right]$.
\\
In this work, we focus solely on vanilla generation methods. We do not discuss other model-specific~\cite{shi2024unchosen}, task-specific ~\cite{yang2021fudge,liu2021dexperts,shi2024lifi}, or meta-generation methods~\cite{welleck2024decoding} (e.g., Tree-of-Thoughts~\cite{yao2024tree}). These specialized decoding approaches are beyond the scope of our current analysis.
\section{Evaluation Benchmarks}
\label{appx:ben}
\subsection{Coding}
HumanEval ~\cite{chen2021humaneval}, MBPP ~\cite{austin2021mbpp} are extensively utilized benchmarks within the measurement of LLM's code generating ability. These benchmarks encompass a vast collection of Python programming problems.
\paragraph{HumanEval} ~\cite{chen2021humaneval} consists of 164 original programming problems by giving docstrings to generate code, which has an average of 9.6 test cases allocated to each problem. We use 0-shot prompt for both unaligned and aligned models.
\paragraph{MBPP} ~\cite{austin2021mbpp} focus on generating code based on textual descriptions, which offers a set of 500 test programming problems, accompanied by three automated test cases per problem. We use 0-shot prompt for aligned models and 3-shot prompt for unaligned models.
\subsection{Math Problem Soving}
We utilize GSM8K ~\cite{cobbe2021gsm8k} for assessing reasoning and problem-solving proficiencies within the domain of mathematics.
\paragraph{GSM8K} ~\cite{cobbe2021gsm8k} collects 1,319 high-quality linguistically diverse grade school math word problems as the test set, and reports 8-shot pass@1 accuracy. We use 0-shot prompt for aligned models and 8-shot prompt for unaligned models.
\subsection{Summarization}
We select the CNN/DailyMail ~\cite{hermann2015CNNDailyMail} and XSUM ~\cite{narayan2018xsum} datasets, which are the most well-studied datasets in the literature on summarization faithfulness. This also ensures domain coverage of news-type data. Importantly, these datasets differ along a central axis studied in summarization:
\paragraph{XSUM} ~\cite{narayan2018xsum} is a dataset with largely abstractive reference summaries (meaning the string overlap between the document and its summary in the dataset is relatively small on average) which feature articles from the British Broadcasting Corporation (BBC). The test splits for the dataset are 11.5K examples. We use 0-shot prompt for aligned models and 1-shot prompt for unaligned models.
\paragraph{CNN/DailyMail} ~\cite{hermann2015CNNDailyMail} is a dataset with largely extractive reference summaries that contain news articles from CNN and the DailyMail along with highlights that act as a summary for the article. The test splits for the dataset are 11.3K examples. We use 0-shot prompt for aligned models and 1-shot prompt for unaligned models.
The model-generated summary is compared against a human-authored reference summary using automated metrics for overall quality ROUGE-L ~\cite{lin2004rouge}.
Note that we randomly select 1,000 cases each from CNNDailyMail and XSUM for evaluation.
\subsection{Translation} 
We evaluate the translation performance on WMT22 ~\cite{kocmi2022wmt22} test sets. 
\paragraph{WMT22 Competition} ~\cite{kocmi2022wmt22} constructed based on more recent content from various domains, including news, social, e-commerce, and conversational domains. The numbers of samples for De $\Rightarrow$ En, En $\Rightarrow$ De, Zh $\Rightarrow$ En and En $\Rightarrow$ Zh tasks are 1984, 2037, 1875 and 2037, respectively. For automatic evaluation, we adopt BLEU ~\cite{papineni2002BLEU} implementated in SacreBLEU ~\cite{post2018SacreBLEU}\footnote{\url{https://github.com/mjpost/sacrebleu}}. We use 3-shot prompt for unaligned models.
\subsection{Commonsense reasoning}
Commonsense reasoning is key for interacting with the world and is still beyond the reach of current natural language understanding systems ~\cite{talmor2019Commonsenseqa}. We consider measuring open-ended performance on two datasets covering a diverse range of commonsense reasoning types from BIG-Bench ~\cite{srivastava2022bigbench}, CommonsenseQA ~\cite{talmor2019Commonsenseqa} and StrategyQA ~\cite{geva2021StrategyQA}.
\paragraph{CommonsenseQA} ~\cite{talmor2019Commonsenseqa} asks commonsense questions about the world involving complex semantics that often require prior knowledge. There are a total of 1.22k instances in the CommonsenseQA validation set. We use 6-shot prompt for aligned models and 1-shot prompt for unaligned models.
\paragraph{StrategyQA}  ~\cite{geva2021StrategyQA} requires models to infer a multi-hop strategy to answer questions. We use the open-domain setting (question-only set) from BIG-Bench ~\cite{srivastava2022bigbench} which contains 2.29k test instances. We use 0-shot prompt for aligned models and 4-shot prompt for unaligned models.
The two BIG-bench tasks do not have training sets, so we select the first ten examples as exemplars in the evaluation set as few-shot exemplars and report accuracy on the rest of the evaluation set.
\subsection{Factual Knowledge}
Factual Knowledge refers to their tendency to generate factual errors. This is considered a critical issue in LLMs because it is challenging for users to identify and poses real-life risks.
\paragraph{FActScore} ~\cite{min2023factscore} scrutinizes the factual accuracy of biographies generated by LLMs for 500 specific individuals. Conducting a pipeline to transform a long-form model generation into pieces of atomic statements and measure the atomic statement's accuracy with retrieved knowledge. We use 0-shot prompt for aligned models.
\subsection{Instruction Following}
For our research, we select the representative broad-coverage benchmark Alpace-eval ~\cite{li2023alpacaeval}. 
\paragraph{Alpace-eval} ~\cite{li2023alpacaeval} assess the LLM's generation quality by 805 prompts from several sources: Vicuna ~\cite{chiang2023vicuna} (80 prompts), Self-instruct ~\cite{zhang2023selfqa} (252 prompts), Open Assistant ~\cite{kopf2023openassistant} (188 prompts), Koala ~\cite{geng2023koala} (156 prompts), HH\_RLHF ~\cite{bai2022hhrhlf} (129 prompts), quantifying the pairwise Win Rate against a reference model, Text-Davinci-003.
\subsection{Open-ended Text Generation}
Open-ended text generation aims to craft fluent and coherent textual continuations of given prompts. Following ~\cite{li2022contrastive}, we evaluate three domains for open-ended text generation: Book, Wikinews, Wikitext.
\paragraph{Book} contains 1,947 prompts collected from BookCorpus ~\cite{Zhubook15}for story generations. We use 0-shot prompt for both unaligned and aligned models.
\paragraph{Wikinews} include 2,000 news articles prompts collected from Wikinews\footnote{http://www.wikinews.org}. We use 0-shot prompt for both unaligned and aligned models.
\paragraph{Wikitext} select 1,314 prompts from wikitext-103 ~\cite{MerityWikitext17} as the Wikipedia representative domain. We use 0-shot prompt for both unaligned and aligned models.
We utilize MAUVE ~\cite{PillutlaMAUVE21} score (the higher the better) to measure the distribution similarity between the set of generated text and the set of gold references.
Note that we randomly select 500 cases each from among the three domains mentioned above for evaluation.
\section{Instruction Template}
\label{sec:instruction_template}
The instruction templates for each dataset are list from Table~\ref{tab:appendix-human-eval-unaligned-prompt} to Table~\ref{tab:appendix-otg-chat-prompt}.
\begin{table*}[h]
    \centering
    \small
    \begin{tabular}{p{\linewidth}}
        \toprule
        \underline{\textbf{\textsc{Prompt for HumanEval}}} \\
        \vspace{-2mm}
        \texttt{\textcolor{blue}{[DOCSTRING]}}\\ 
        \bottomrule
    \end{tabular}
    \caption{0-shot prompt for HumanEval (unaligned model).}
    \label{tab:appendix-human-eval-unaligned-prompt}
\end{table*}

\begin{table*}[h]
    \centering
    \small
    \begin{tabular}{p{\linewidth}}
        \toprule
        \underline{\textbf{\textsc{Prompt for HumanEval}}} \\
        \vspace{-2mm}
        Please complete the remaining Python function code based on the following docstring content. \\
        \texttt{\textcolor{blue}{[DOCSTRING]}}\\ 
        \bottomrule
    \end{tabular}
    \caption{0-shot prompt for HumanEval (aligned model).}
    \label{tab:appendix-human-eval-chat-prompt}
\end{table*}
\begin{table*}[h]
    \centering
    \small
    \begin{tabular}{p{0.95\linewidth}}
        \toprule
        \underline{\textbf{\textsc{Prompt for MBPP}}} \\
        \vspace{-2mm}
        You are an expert Python programmer, and here is your task: Write a function to find the similar elements from the given two tuple lists. Your code should pass these tests: \\
        assert similar\_elements((3, 4, 5, 6),(5, 7, 4, 10)) == (4, 5) \\
        assert similar\_elements((1, 2, 3, 4),(5, 4, 3, 7)) == (3, 4) \\
        assert similar\_elements((11, 12, 14, 13),(17, 15, 14, 13)) == (13, 14) \\
        \texttt{[BEGIN]} \\
        def similar\_elements(test\_tup1, test\_tup2):\\
        \hspace{4mm} res = tuple(set(test\_tup1) \& set(test\_tup2))\\
        \hspace{4mm} return (res) \\
        \texttt{[DONE]} \\
        \vspace{2mm}
        You are an expert Python programmer, and here is your task: Write a python function to identify non-prime numbers. Your code should pass these tests: \\
        assert is\_not\_prime(2) == False\\
        assert is\_not\_prime(10) == True\\
        assert is\_not\_prime(35) == True\\
        \texttt{[BEGIN]} \\
        import math\\
        def is\_not\_prime(n):\\
        \hspace{4mm} result = False\\
        \hspace{4mm} for i in range(2,int(math.sqrt(n)) + 1):\\
        \hspace{8mm} if n \% i == 0:\\
        \hspace{12mm} result = True\\
        \hspace{4mm} return result\\
        \texttt{[DONE]} \\
        \vspace{2mm}
        You are an expert Python programmer, and here is your task: Write a function to find squares of individual elements in a list using lambda function. Your code should pass these tests: \\
        assert square\_nums(\texttt{[}1, 2, 3, 4, 5, 6, 7, 8, 9, 10\texttt{]})==\texttt{[}1, 4, 9, 16, 25, 36, 49, 64, 81, 100\texttt{]})\\
        assert square\_nums(\texttt{[}10,20,30\texttt{]}))==(\texttt{[}100,400,900\texttt{]}))\\
        assert square\_nums(\texttt{[}12,15\texttt{]}))==(\texttt{[}144,225\texttt{]}))\\
        \texttt{[BEGIN]} \\
        def square\_nums(nums):\\
        \hspace{4mm} square\_nums = list(map(lambda x: x ** 2, nums)) \\
        \hspace{4mm} return square\_nums\\
        \texttt{[DONE]} \\
        \vspace{2mm}
        You are an expert Python programmer, and here is your task: \texttt{\textcolor{blue}{[TASK\_DEFINATION]}}. Your code should pass these tests: \\
        \texttt{\textcolor{blue}{[TEST\_CASE\_1]}}\\
        \texttt{\textcolor{blue}{[TEST\_CASE\_2]}}\\
        \texttt{\textcolor{blue}{[TEST\_CASE\_2]}}\\
        \texttt{[BEGIN]} \\
        \bottomrule
    \end{tabular}
    \caption{3-shot promp for MBPP (unaligned model).}
    \label{tab:appendix-MBPP-unaligned-prompt}
\end{table*}

\begin{table*}[h]
    \centering
    \small
    \begin{tabular}{p{0.95\linewidth}}
        \toprule
        \underline{\textbf{\textsc{Prompt for MBPP}}} \\
        \vspace{-2mm}
        You are an expert Python programmer, and here is your task: \texttt{\textcolor{blue}{[TASK\_DEFINATION]}}. Your code should pass these tests: \\
        \texttt{\textcolor{blue}{[TEST\_CASE\_1]}}\\
        \texttt{\textcolor{blue}{[TEST\_CASE\_2]}}\\
        \texttt{\textcolor{blue}{[TEST\_CASE\_2]}}\\
        \bottomrule
    \end{tabular}
    \caption{0-shot promp for MBPP (aligned model).}
    \label{tab:appendix-MBPP-chat-prompt}
\end{table*}
\begingroup
\begin{table*}[h]
    \centering
    \small
    \begin{tabular}{p{0.95\linewidth}}
        \toprule
        \underline{\textbf{\textsc{Prompt for GSM8K}}} \\
        \vspace{-2mm}
        \textbf{Question:} There are 15 trees in the grove. Grove workers will plant trees in the grove today. After they are done, there will be 21 trees. How many trees did the grove workers plant today? \\
        \vspace{-1mm}
        \textbf{Answer:} There are 15 trees originally. Then there were 21 trees after some more were planted. So there must have been 21 - 15 = 6. The answer is 6. \\
        \vspace{2mm}
        \textbf{Question:} If there are 3 cars in the parking lot and 2 more cars arrive, how many cars are in the parking lot? \\
        \vspace{-1mm}
        \textbf{Answer:} There are originally 3 cars. 2 more cars arrive. 3 + 2 = 5. The answer is 5. \\
        \vspace{2mm}
        \textbf{Question:} Leah had 32 chocolates and her sister had 42. If they ate 35, how many pieces do they have left in total? \\
        \vspace{-1mm}
        \textbf{Answer:} Originally, Leah had 32 chocolates. Her sister had 42. So in total they had 32 + 42 = 74. After eating 35, they had 74 - 35 = 39. The answer is 39. \\
        \vspace{2mm}
        \textbf{Question:} Question: Jason had 20 lollipops. He gave Denny some lollipops. Now Jason has 12 lollipops. How many lollipops did Jason give to Denny?\\
        \vspace{-1mm}
        \textbf{Answer:} Jason started with 20 lollipops. Then he had 12 after giving some to Denny. So he gave Denny 20 - 12 = 8. The answer is 8. \\
        \vspace{2mm}
        \textbf{Question:} Shawn has five toys. For Christmas, he got two toys each from his mom and dad. How many toys does he have now?\\
        \vspace{-1mm}
        \textbf{Answer:} Shawn started with 5 toys. If he got 2 toys each from his mom and dad, then that is 4 more toys. 5 + 4 = 9. The answer is 9. \\
        \vspace{2mm}
        \textbf{Question:} There were nine computers in the server room. Five more computers were installed each day, from monday to thursday. How many computers are now in the server room?\\
        \vspace{-1mm}
        \textbf{Answer:} There were originally 9 computers. For each of 4 days, 5 more computers were added. So 5 * 4 = 20 computers were added. 9 + 20 is 29. The answer is 29. \\
        \vspace{2mm}
        \textbf{Question:} Michael had 58 golf balls. On tuesday, he lost 23 golf balls. On wednesday, he lost 2 more. How many golf balls did he have at the end of wednesday? \\
        \vspace{-1mm}
        \textbf{Answer:} Michael started with 58 golf balls. After losing 23 on tuesday, he had 58 - 23 = 35. After losing 2 more, he had 35 - 2 = 33 golf balls. The answer is 33. \\
        \vspace{2mm}
        \textbf{Question:} Olivia has \$23. She bought five bagels for \$3 each. How much money does she have left? \\
        \vspace{-1mm}
        \textbf{Answer:} Olivia had 23 dollars. 5 bagels for 3 dollars each will be 5 x 3 = 15 dollars. So she has 23 - 15 dollars left. 23 - 15 is 8. The answer is 8. \\
        \vspace{2mm}
        \textbf{Question:} \texttt{\textcolor{blue}{[QUESTION]}}\\
        \vspace{-1mm}
        \textbf{Answer:} \\
        \bottomrule
    \end{tabular}
    \caption{    
    8-shot prompt for GSM8K (unaligned model).
    }
    \label{tab:appendix-gsm8k-unaligned-prompt}
\end{table*}
\endgroup

\begingroup
\begin{table*}[h]
    \centering
    \small
    \begin{tabular}{p{0.95\linewidth}}
        \toprule
        \underline{\textbf{\textsc{Prompt for GSM8K}}} \\
        \vspace{-2mm}
        Please answer the math questions below.\\
        \texttt{\textcolor{blue}{[QUESTION]}} \\
        You need to first take step-by-step reasoning and then give the final result.\\
        \bottomrule
    \end{tabular}
    \caption{    
    0-shot prompt for GSM8K (aligned model).
    }
    \label{tab:appendix-gsm8k-chat-prompt}
\end{table*}
\endgroup
\begin{table*}[h]
    \centering
    \small
    \begin{tabular}{p{0.95\linewidth}}
        \toprule
        \underline{\textbf{\textsc{Prompt for Xsum}}} \\
        \vspace{-2mm}
        \textbf{Article:} The Bath-born player, 28, has made 36 appearances for the Dragons since joining from Wasps in 2015. He is in his second season and signed a contract extension in December 2016. Dragons forwards coach Ceri Jones said: "It's a big blow. Eddie has been excellent all year for us, he has really stepped up to the mark and will be a big loss." However, Jones says Jackson's misfortune can be a chance for others to thrive. "We are very fortunate to have the likes of Ollie Griffiths, Harrison Keddie, James Thomas who can come into the back-row," said Jackson. "Harri has shown glimpses of what he can do all season and there's definitely a player there, so this is an opportunity." Dragons travel to Munster in the Pro12 on Friday.\\
        \vspace{-1mm}
        Summarize the above article in 1 sentence. \\
        Newport Gwent Dragons number eight Ed Jackson has undergone shoulder surgery and faces a spell on the sidelines.
        \vspace{2mm}
        \textbf{Article:} \texttt{\textcolor{blue}{[ARTICLE]}}\\
        SSummarize the above article in 1 sentence.\\ 
        \bottomrule
    \end{tabular}
    \caption{1-shot prompt for XSUM (unaligned model).}
    \label{tab:appendix-xsum-chat-prompt}
\end{table*}

\begin{table*}[h]
    \centering
    \small
    \begin{tabular}{p{0.95\linewidth}}
        \toprule
        \underline{\textbf{\textsc{Prompt for Xsum}}} \\
        \vspace{-2mm}
        \textbf{Article:} \texttt{\textcolor{blue}{[ARTICLE]}}\\ \\Summarize the above article in 1 sentence.\\ 
        \bottomrule
    \end{tabular}
    \caption{0-shot prompt for XSUM (aligned model).}
    \label{tab:appendix-xsum-nochat-prompt}
\end{table*}
\begin{table*}[h]
    \centering
    \small
    \begin{tabular}{p{0.95\linewidth}}
        \toprule
        \underline{\textbf{\textsc{Prompt for CNNDailymail}}} \\
        \vspace{-2mm}
        \textbf{Article:} PARIS, France (CNN) -- Interpol on Monday took the unprecendented step of making a global appeal for help to identify a man from digitally reconstructed photos taken from the Internet that it said showed him sexually abusing underage boys. This moving image shows how police used software to unscramble the image. (Source: Interpol) The man's face was disguised by digital alteration, but the images were capable of being restored, according to a bulletin from Interpol -- the international police agency based in Lyon, France. Interpol Secretary General Ronald K. Noble said the pictures have been on the the Internet for several years, but investigators have been unable to determine the man's identity or nationality. "We have tried all other means to identify and to bring him to justice, but we are now convinced that without the public's help this sexual predator could continue to rape and sexually abuse young children whose ages appear to range from six to early teens," Noble said. He said there is "very good reason to believe that he travels the world in order to sexually abuse and exploit vulnerable children." Interpol has determined the photos were taken in Vietnam and Cambodia. "The decision to make public this man's picture was not one which was taken lightly," said Kristin Kvigne, assistant director of Interpol's Trafficking in Human Beings Unit. The suspect's photo and more information can be seen online at Interpol's Web site. E-mail to a friend .\\
        Summarize the above article in 3 sentences. \\
        Man posted photos on the Internet of himself sexually abusing underage boys . Computer experts managed to undo digital masking to reveal the man . Man abused 12 boys in Vietnam and Cambodia . \\
        \vspace{2mm}
        \textbf{Article:} \texttt{\textcolor{blue}{[ARTICLE]}}\\

        Summarize the above article in 3 sentences.\\ 
        \bottomrule
    \end{tabular}
    \caption{1-shot prompt for CNN/Dailymail (unaligned model).}
    \label{tab:appendix-cnndaily-unaligned-prompt}
\end{table*}

\begin{table*}[h]
    \centering
    \small
    \begin{tabular}{p{0.95\linewidth}}
        \toprule
        \underline{\textbf{\textsc{Prompt for CNNDailymail}}} \\
        \vspace{-2mm}
        \textbf{Article:} \texttt{\textcolor{blue}{[ARTICLE]}}\\ \\Summarize the above article in 3 sentences.\\ 
        \bottomrule
    \end{tabular}
    \caption{0-shot prompt for CNN/Dailymail (aligned model).}
    \label{tab:appendix-cnndaily-chat-prompt}
\end{table*}
\begingroup
\begin{table*}[h]
    \centering
    \small
    \begin{tabular}{p{0.95\linewidth}}
        \toprule
        \underline{\textbf{\textsc{Prompt for WMT De$\Rightarrow$En }}} \\
        \vspace{-2mm}
        Translate the following sentence from German to English.\\
        \vspace{-1mm}
        \texttt{[}GERMAN\texttt{]} Frau Schroedter, ich bin gerne bereit, die damit zusammenhängenden Fakten zu prüfen, wenn mir Ihr Brief vorliegt. \\
        \texttt{[}ENGLISH\texttt{]} Yes, Mrs Schroedter, I shall be pleased to look into the facts of this case when I have received your letter.\\
        \vspace{2mm}
        Translate the following sentence from German to English.\\
        \vspace{-1mm}
        \texttt{[}GERMAN\texttt{]} Das ist der Fall von Alexander Nikitin. \\
        \texttt{[}ENGLISH\texttt{]} It is the case of Alexander Nikitin. \\
        \vspace{2mm}
        Translate the following sentence from German to English.\\
        \vspace{-1mm}
        \texttt{[}GERMAN\texttt{]} Meine Frage betrifft eine Angelegenheit, die am Donnerstag zur Sprache kommen wird und auf die ich dann erneut verweisen werde. \\
        \texttt{[}ENGLISH\texttt{]} My question relates to something that will come up on Thursday and which I will then raise again. \\
        \vspace{2mm}
        Translate the following sentence from German to English.\\
        \vspace{-1mm}
        \texttt{[}GERMAN\texttt{]} \texttt{\textcolor{blue}{[GERMAN\_TEXT]}} \\
        \texttt{[}ENGLISH\texttt{]}  \\
        \bottomrule
    \end{tabular}
    \caption{3-shot prompt for WMT De$\Rightarrow$En (unaligned model).}
    \label{tab:appendix-wmtde2en-prompt}
\end{table*}
\endgroup

\begingroup
\begin{table*}[h]
    \centering
    \small
    \begin{tabular}{p{0.95\linewidth}}
        \toprule
        \underline{\textbf{\textsc{Prompt for WMT En$\Rightarrow$De}}} \\
        \vspace{-2mm}
        Translate the following sentence from English to German.\\
        \vspace{-1mm}
        \texttt{[}ENGLISH\texttt{]} Yes, Mrs Schroedter, I shall be pleased to look into the facts of this case when I have received your letter.\\
        \texttt{[}GERMAN\texttt{]} Frau Schroedter, ich bin gerne bereit, die damit zusammenhängenden Fakten zu prüfen, wenn mir Ihr Brief vorliegt. \\
        \vspace{2mm}
        Translate the following sentence from English to German.\\
        \vspace{-1mm}
        \texttt{[}ENGLISH\texttt{]} It is the case of Alexander Nikitin. \\
        \texttt{[}GERMAN\texttt{]} Das ist der Fall von Alexander Nikitin. \\
        \vspace{2mm}
        Translate the following sentence from English to German.\\
        \vspace{-1mm}
        \texttt{[}GERMAN\texttt{]} Meine Frage betrifft eine Angelegenheit, die am Donnerstag zur Sprache kommen wird und auf die ich dann erneut verweisen werde. \\
        \texttt{[}ENGLISH\texttt{]} My question relates to something that will come up on Thursday and which I will then raise again. \\
        \vspace{2mm}
        Translate the following sentence from English to German.\\
        \vspace{-1mm}
        \texttt{[}ENGLISH\texttt{]} \texttt{\textcolor{blue}{[ENGLISH\_TEXT]}}\\
        \texttt{[}GERMAN\texttt{]} \\
        \bottomrule
    \end{tabular}
    \caption{    
    3-shot prompt for WMT En$\Rightarrow$De (unaligned model).
    }
    \label{tab:appendix-wmten2de-prompt}
\end{table*}
\endgroup

\begingroup
\begin{table*}[h]
    \centering
    \small
    \begin{tabular}{p{0.95\linewidth}}
        \toprule
        \underline{\textbf{\textsc{Prompt for WMT Zh$\Rightarrow$En}}} \\
        \vspace{-2mm}
        Translate the following sentence from Chinses to English.\\
        \vspace{-1mm}
        \texttt{[}CHINESE\texttt{]} \begin{CJK}{UTF8}{gbsn}柏林 — — 2008年爆发的全球金融和经济危机是自大萧条以来最严峻的一次经济压力测试，也是自二战以来社会和政治制度所面临的最严重挑战。\end{CJK} \\
        \texttt{[}ENGLISH\texttt{]} BERLIN – The global financial and economic crisis that began in 2008 was the greatest economic stress-test since the Great Depression, and the greatest challenge to social and political systems since World War II. \\
        \vspace{2mm}
        Translate the following sentence from Chinses to English.\\
        \vspace{-1mm}
        \texttt{[}CHINESE\texttt{]} \begin{CJK}{UTF8}{gbsn}欧洲在避免债务和捍卫欧元的名义下正变得谨慎，而美国已经在许多方面行动起来，以利用这一理想的时机来实行急需的结构性改革。\end{CJK} \\
        \texttt{[}ENGLISH\texttt{]} Europe is being cautious in the name of avoiding debt and defending the euro, whereas the US has moved on many fronts in order not to waste an ideal opportunity to implement badly needed structural reforms. \\
        \vspace{2mm}
        Translate the following sentence from Chinses to English.\\
        \vspace{-1mm}
        \texttt{[}CHINESE\texttt{]} \begin{CJK}{UTF8}{gbsn}百年愚顽\end{CJK} \\
        \texttt{[}ENGLISH\texttt{]} One Hundred Years of Ineptitude \\
        \vspace{2mm}
        Translate the following sentence from Chinses to English.\\
        \vspace{-1mm}
        \texttt{[}CHINESE\texttt{]} \texttt{\textcolor{blue}{[CHINESE\_TEXT]}}\\
        \texttt{[}ENGLISH\texttt{]} \\
        \bottomrule
    \end{tabular}
    \caption{    
    3-shot prompt for WMT Zh$\Rightarrow$En (unaligned model).
    }
    \label{tab:appendix-wmtzh2en-prompt}
\end{table*}
\endgroup
\begingroup
\begin{table*}[h]
    \centering
    \small
    \begin{tabular}{p{0.95\linewidth}}
        \toprule
        \underline{\textbf{\textsc{Prompt for WMT En$\Rightarrow$Zh}}} \\
        \vspace{-2mm}
        Translate the following sentence from English to Chinese.\\
        \vspace{-1mm}
        \texttt{[}ENGLISH\texttt{]} BERLIN – The global financial and economic crisis that began in 2008 was the greatest economic stress-test since the Great Depression, and the greatest challenge to social and political systems since World War II. \\
        \texttt{[}CHINESE\texttt{]} \begin{CJK}{UTF8}{gbsn}柏林 — — 2008年爆发的全球金融和经济危机是自大萧条以来最严峻的一次经济压力测试，也是自二战以来社会和政治制度所面临的最严重挑战。\end{CJK} \\
        \vspace{2mm}
        Translate the following sentence from English to Chinese.\\
        \vspace{-1mm}
        \texttt{[}ENGLISH\texttt{]} Europe is being cautious in the name of avoiding debt and defending the euro, whereas the US has moved on many fronts in order not to waste an ideal opportunity to implement badly needed structural reforms. \\
        \texttt{[}CHINESE\texttt{]} \begin{CJK}{UTF8}{gbsn}欧洲在避免债务和捍卫欧元的名义下正变得谨慎，而美国已经在许多方面行动起来，以利用这一理想的时机来实行急需的结构性改革。\end{CJK} \\
        \vspace{2mm}
        Translate the following sentence from English to Chinese.\\
        \vspace{-1mm}
        \texttt{[}ENGLISH\texttt{]} One Hundred Years of Ineptitude \\
        \texttt{[}CHINESE\texttt{]} \begin{CJK}{UTF8}{gbsn}百年愚顽\end{CJK} \\
        \vspace{2mm}
        Translate the following sentence from English to Chinese.\\
        \vspace{-1mm}
        \texttt{[}ENGLISH\texttt{]} \texttt{\textcolor{blue}{[ENGLISH\_TEXT]}}\\
        \texttt{[}CHINESE\texttt{]} \\
        \bottomrule
    \end{tabular}
    \caption{3-shot prompt for WMT En$\Rightarrow$Zh (unaligned model).}
    \label{tab:appendix-wmten2zh-prompt}
\end{table*}
\endgroup
\begingroup
\begin{table*}[h]
    \centering
    \small
    \begin{tabular}{p{0.95\linewidth}}
        \toprule
        \underline{\textbf{\textsc{Prompt for CommonsenseQA}}} \\
        \vspace{-2mm}
        \textbf{Question:} What do people use to absorb extra ink from a fountain pen? Answer Choices: (a) shirt pocket (b) calligrapher's hand (c) inkwell (d) desk drawer (e) blotter \\
        \vspace{-1mm}
        \textbf{Answer:} The answer must be an item that can absorb ink. Of the above choices, only blotters are used to absorb ink. So the answer is (e). \\
        \vspace{2mm}
        \textbf{Question:} What home entertainment equipment requires cable? \\
        Answer Choices: (a) radio shack (b) substation (c) television (d) cabinet \\
        \vspace{-1mm}
        \textbf{Answer:} The answer must require cable. Of the above choices, only television requires cable. So the answer is (c). \\
        \vspace{2mm}
        \textbf{Question:} The fox walked from the city into the forest, what was it looking for? Answer Choices: (a) pretty flowers (b) hen house (c) natural habitat (d) storybook\\
        \vspace{-1mm}
        \textbf{Answer:} The answer must be something in the forest. Of the above choices, only natural habitat is in the forest.  So the answer is (b). \\
        \vspace{2mm}
        \textbf{Question:} Sammy wanted to go to where the people were. Where might he go? Answer Choices: (a) populated areas (b) race track (c) desert (d) apartment (e) roadblock \\
        \vspace{-1mm}
        \textbf{Answer:} The answer must be a place with a lot of people. Of the above choices, only populated areas have a lot of people. So the answer is (a). \\
        \vspace{2mm}
        \textbf{Question:} Where do you put your grapes just before checking out? Answer Choices: (a) mouth (b) grocery cart (c)super market (d) fruit basket (e) fruit market\\
        \vspace{-1mm}
        \textbf{Answer:} The answer should be the place where grocery items are placed before checking out. Of the above choices, grocery cart makes the most sense for holding grocery items. So the answer is (b). \\
        \vspace{2mm}
        \textbf{Question:} Google Maps and other highway and street GPS services have replaced what? Answer Choices: (a) united states (b) mexico (c) countryside (d) atlas \\
        \vspace{-1mm}
        \textbf{Answer:} The answer must be something that used to do what Google Maps and GPS services do, which is to give directions. Of the above choices, only atlases are used to give directions. So the answer is (d). \\
        \textbf{Question:} \texttt{\textcolor{blue}{[QUESTION]}} \\
        \vspace{-1mm}
        \textbf{Answer:} \\
        \bottomrule
    \end{tabular}
    \caption{6-shot prompt for CommonsenseQA (unaligned model).}
    \label{tab:appendix-commonsenseqa-unaligned-prompt}
\end{table*}
\endgroup

\begingroup
\begin{table*}[h]
    \centering
    \small
    \begin{tabular}{p{0.95\linewidth}}
        \toprule
        \underline{\textbf{\textsc{Prompt for CommonsenseQA}}} \\
        \vspace{-2mm}
        Which choice is the correct answer to the question?\\
        \textbf{Question:} \texttt{\textcolor{blue}{[QUESTION]}} \\
        \vspace{-1mm}
        \textbf{Answer:} The answer must be an item that can absorb ink. Of the above choices, only blotters are used to absorb ink. So the answer is (e). \\
        Let's think step by step. \\
        \bottomrule
    \end{tabular}
    \caption{0-shot prompt for CommonsenseQA (aligned model).}
    \label{tab:appendix-commonsenseqa-chat-prompt}
\end{table*}
\endgroup
\begin{table*}[h]
    \centering
    \small
    \begin{tabular}{p{0.95\linewidth}}
        \toprule
        \underline{\textbf{\textsc{Prompt for StrategyQA}}} \\
        \vspace{-2mm}
        \textbf{Question:} Do hamsters provide food for any animals? \\
        \vspace{-1mm}
        \textbf{Answer:} Hamsters are prey animals. Prey are food for predators. Thus, hamsters provide food for some animals. So the answer is yes. \\
        \vspace{2mm}
        \textbf{Question:} Could Brooke Shields succeed at University of Pennsylvania? \\
        \vspace{-1mm}
        \textbf{Answer:} Brooke Shields went to Princeton University. Princeton University is about as academically rigorous as the University of Pennsylvania. Thus, Brooke Shields could also succeed at the University of Pennsylvania.  So the answer is yes. \\
        \vspace{2mm}
        \textbf{Question:} Yes or no: Hydrogen's atomic number squared exceeds number of Spice Girls? \\
        \vspace{-1mm}
        \textbf{Answer:} Hydrogen has an atomic number of 1. 1 squared is 1. There are 5 Spice Girls. Thus, Hydrogen's atomic number squared is less than 5. So the answer is no. \\
        \vspace{2mm}
        \textbf{Question:} Yes or no: Is it common to see frost during some college commencements? \\
        \vspace{-1mm}
        \textbf{Answer:} College commencement ceremonies can happen in December, May, and June. December is in the winter, so there can be frost. Thus, there could be frost at some commencements. So the answer is yes. \\
        \vspace{2mm}
        \textbf{Question:} \texttt{\textcolor{blue}{[QUESTION]}} \\
        \vspace{-1mm}
        \textbf{Answer:} \\
        \bottomrule
    \end{tabular}
    \caption{4-shot prompt for StrategyQA (unaligned model).}
    \label{tab:appendix-strategyqa-unaligned-prompt}
\end{table*}

\begin{table*}[h]
    \centering
    \small
    \begin{tabular}{p{0.95\linewidth}}
        \toprule
        \underline{\textbf{\textsc{Prompt for CommonsenseQA}}} \\
        \vspace{-2mm}
        Which choice is the correct answer to the question?\\
        \textbf{Question:} \texttt{\textcolor{blue}{[QUESTION]}} \\
        \vspace{-1mm}
        \textbf{Answer:} The answer must be an item that can absorb ink. Of the above choices, only blotters are used to absorb ink. So the answer is (e). \\
        Let's think step by step. \\
        \bottomrule
    \end{tabular}
    \caption{0-shot prompt for StrategyQA (aligned model).}
    \label{tab:appendix-strategyqa-chat-prompt}
\end{table*}
\begin{table*}[h]
    \centering
    \small
    \begin{tabular}{p{0.95\linewidth}}
        \toprule
        \underline{\textbf{\textsc{Prompt for Book, WikiNews and WikiText}}} \\
        \vspace{-2mm}
        \texttt{\textcolor{blue}{[BEGIN\_OF\_TEXT]}}\\ 
        \bottomrule
    \end{tabular}
    \caption{0-shot prompt for Book, Wikinews and Wikitext (unaligned model).}
    \label{tab:appendix-otg-unaligned-prompt}
\end{table*}

\begin{table*}[h]
    \centering
    \small
    \begin{tabular}{p{0.95\linewidth}}
        \toprule
        \underline{\textbf{\textsc{Prompt for Book, WikiNews and WikiText}}} \\
        \vspace{-2mm}
        Please help me complete the text continuation based on the following content. \\
        \texttt{\textcolor{blue}{[BEGIN\_OF\_TEXT]}}\\ 
        \bottomrule
    \end{tabular}
    \caption{0-shot prompt for Book, Wikinews and Wikitext (aligned model).}
    \label{tab:appendix-otg-chat-prompt}
\end{table*}
\section{Different Foundation Models}
\label{sec:foundation}
\begin{table*}[t]
    \small
    \centering
    \resizebox{\textwidth}{!}{
    \begin{tabular}{c|c|cccccc|cccccc}
    \toprule
    \addlinespace[-0.0001ex]
    \multirow{2}{*}{\textbf{Model}}& \multirow{2}{*}{\textbf{Dataset}}& \multicolumn{6}{c|}{\textbf{Deterministic Methods}}& \multicolumn{6}{c}{\textbf{Stochastic Methods}} \\
    \cline{3-14}
    \addlinespace[0.2ex]
    &&\textbf{Greedy} & \textbf{BS} & \textbf{DBS} &\textbf{CS} & \textbf{FSD} &\textbf{FSD-d}  &\textbf{Temp} &\textbf{Top-$p$} &\textbf{Top-$k$} &\textbf{$\eta$} &\textbf{Miro} & \textbf{Typical} \\
    \cline{1-14}
    \addlinespace[0.2ex]
    \multirow{3}{*}{CodeLlama-7b}&MBPP& 35.40  &34.20  &35.00  &\redthree 36.00  &\redtwo 37.00  &\redone 39.60  &35.00  &32.80  &\greenthree 25.40  &\greentwo 23.60  &\greenone 21.20  &31.80 \\
    &GSM8K & 11.98  &\redthree 13.12  &12.21  &12.89  &\redtwo 13.50  &\redone 13.80  &12.59  &12.66  &\greenthree 8.64  &\greentwo 7.43  &\greenone 6.90  &8.87 \\
    &Wikinews &\greenthree 10.49  &\greentwo 9.81  &\greenone 8.85  &87.99  &\redtwo 97.19  &94.35  &90.89  &94.63  &\redthree 96.69  &\redone 97.59  &96.34  &94.06 \\
    \cline{1-14}     \addlinespace[0.2ex]
    \multirow{3}{*}{CodeLlama-7b-Instruct}&MBPP& 36.80  &\redtwo 40.80  &\redone 41.60  &37.00  &37.20  &36.60  &\redthree 39.00  &37.60  &\greenthree 35.60  &\greentwo 35.40  &\greenone 34.40  &38.20 \\
    &GSM8K &22.14  &\redtwo 27.75  &\redone 28.35  &23.28  &22.67  &21.91  &\redthree 23.96  &22.14  &\greenone 18.04  &\greenthree 19.26  &\greentwo 18.27  &21.99 \\
    &Wikinews & 90.11  &\redone 96.75  &\greentwo 85.83  &90.39  &\redthree 92.87  &92.21  &\greenthree 87.69  &89.26  &90.39  &90.31  &\redtwo 94.32  &\greenone 84.99  \\
    \cline{1-14}     \addlinespace[0.2ex]
    \multirow{3}{*}{Qwen-7B}&MBPP& 33.00  &\redone 34.40  &33.20  &28.40  &33.00  &\redthree 33.60  &\redtwo 33.80  &27.40  &\greentwo 19.80  &\greenthree 25.80  &\greenone 18.40  &27.00 \\
    &GSM8K & 53.22  &\redone 57.32  &\redtwo 56.56  &50.04  &53.53  &\redthree 54.59  &53.90  &\greenthree 47.46  &\greentwo 38.67  &49.05  &\greenone 36.24  &51.86 \\
    &Wikinews & \greenone 50.58  &\greenthree 61.66  &\greentwo 51.59  &94.50  &94.22  &\redthree 95.24  &94.50  &94.94  &\redtwo 95.51  &\redone 96.08  &93.94  &94.69 \\
    \cline{1-14}     \addlinespace[0.2ex]
     \multirow{3}{*}{Qwen-7B-Chat}&MBPP&30.40  &\redtwo 30.80  &\redone 33.60  &\greenthree 25.80  &\redtwo 30.80  &29.80  &30.00  &28.80  &26.80  &\greenone 24.20  &\greentwo 25.00  &27.20 \\
    &GSM8K & 48.29  &\redone 51.48  &\redtwo 51.18  &43.82  &47.46  &48.37  &\redthree 48.52  &45.34  &\greenone 41.02  &\greenthree 43.37  &\greentwo 41.85  &43.67 \\
    &Wikinews & \greenone 73.43  &89.12  &\redtwo 90.75  &\redone 91.87  &89.40  &88.07  &\redthree 89.85  &88.11  &85.43  &84.37  &\greenthree 81.37  &\greentwo 80.04 \\
    \cline{1-14}     \addlinespace[0.2ex]
    \multirow{3}{*}{MPT-7b}&MBPP& 18.20  &\redone 22.80  &21.00  &21.20  &\redthree 21.40  &\redtwo 21.80  &19.00  &14.80  &\greenthree 11.20  &\greentwo 8.40  &\greenone 6.60  &11.40 \\
    &GSM8K & 8.64  &9.63  &\redtwo 9.70  &\redone 10.24  &8.95  &\redone 10.24  &9.55  &6.75  &6.37  &\greenthree 5.91  &\greenone 5.08  &\greentwo 5.76 \\
    &Wikinews &  \greenthree 22.44  &\greenone 6.08  &\greentwo 7.30  &87.85  &\redtwo 97.96  &97.58  &96.89  &96.19  &\redone 98.56  &96.95  &\redthree 97.83  &97.80  \\
    \cline{1-14}     \addlinespace[0.2ex]
    \multirow{3}{*}{MPT-7b-Instruct}&MBPP&20.80  &\redone 25.40  &\redtwo 23.80  &23.20  &23.20  &22.20  &\redthree 23.40  &18.60  &\greentwo 15.80  &\greenone 14.20  &\greenone 14.20  &16.60 \\
    &GSM8K & 4.93  &\greenthree 2.88  &4.09  &\greenthree 2.88  &\redone 6.75  &\redtwo 5.91  &\redone 6.75  &4.85  &5.46  &\greentwo 2.65  &4.09  &\greenone 2.58 \\
    &Wikinews & 92.76  &94.88  &\greenthree 87.21 &\redone 97.10  &93.19  & 96.11  &95.53  &96.04  &\redtwo 97.02  &\greenone 50.22  &\redthree 96.53  &\greentwo 53.60  \\
    \cline{1-14}     \addlinespace[0.2ex]
    \multirow{3}{*}{Mistral-7B}&MBPP&  36.40&  \redone41.80&  \redtwo41.60&  \redthree39.60&  39.20&  38.60&  37.40&  32.60&  \greenthree28.00&  \greentwo24.20&  \greenone21.80& 28.40  \\
    &GSM8K & 43.90 & \redone46.70& \redthree45.79& 43.44& 45.26& \redtwo45.94& 43.75& 38.81& 38.58&\greentwo28.20& \greenone23.58& \redthree35.19 \\
    &Wikinews & \greenthree46.74& \greentwo45.81& \greenone35.44& 93.88& 92.58& 89.67& 88.60& \redone94.70& 91.64& 92.78& \redtwo94.03& \redthree93.34 \\
   \cline{1-14}     \addlinespace[0.2ex]
    \multirow{3}{*}{Mistral-7B-Instruct}&MBPP& \redone29.00 & 28.20& 27.20& 27.40& 27.60& 27.40& \redtwo28.80& \redthree28.40& 27.40& \greenthree26.60& \greenone25.80& \greentwo26.40 \\
    &GSM8K & 43.75& \redone49.05& \redtwo46.93& \greenone42.61& 43.52& 43.67& \redthree44.05& 43.82& 43.59& \greentwo43.29& \greenthree43.52& 44.43 \\
    &Wikinews & 79.89& 82.40& 83.97& 76.86& 85.96& 83.47& \redthree88.36 & \redthree89.45& 83.78& 84.85& 82.54& \redone90.61 \\
   \cline{1-14}     \addlinespace[0.2ex]
    \multirow{3}{*}{deepseek-moe-16b-base}&MBPP&35.20 & \redthree36.20 & 35.80 & 34.20 & \redtwo36.60 & \redone36.80 & 28.20 & 26.60 & \greentwo19.00 & \greenone18.80 & \greenthree21.20 & 28.20 \\
    &GSM8K & \redtwo18.95 & 18.20 & \redthree18.42 & \redone19.94 & \redthree18.42 & 18.27 & 16.3  & 12.13 & 12.89 & 12.66               & 11.98 & 10.77 \\
    &Wikinews & \greentwo41.16&\greenthree42.80&\greenone40.66&94.79&95.42&\redthree96.74&96.30&\redtwo97.42&\redone98.01&96.83&96.16&95.69\\
    \cline{1-14}     \addlinespace[0.2ex]
     \multirow{3}{*}{deepseek-moe-16b-chat}&MBPP& \redthree41.00 & \redone41.80 & \redtwo41.20 & 39.20 & 39.40 & 38.20 & 36.20 & 36.40 & \greentwo31.20 & 33.20 & \greenone31.00 & \redthree32.20 \\
    &GSM8K & \redthree50.11 & \redtwo50.64 & 48.90 & \redone50.95 & 46.25 & 47.61 & 39.50 & 45.41 & 36.77 & \greentwo31.69 & \greenone27.14 & \greenthree36.09 \\
    &Wikinews & \greenone75.44 &\greentwo80.34&87.94&\redtwo92.38 &\greenthree83.90&91.74&90.07&89.35&\redone92.41&91.04&\redthree91.68&90.44\\
    \cline{1-14}     \addlinespace[0.2ex]
     \multirow{3}{*}{Llama-3-8B}&MBPP& 43.20 & \redtwo 49.80 & \redone 52.60 & 45.60 & 48.20 & \redthree 49.00 & 43.20 & \greenone 26.00 & \greentwo 26.80 & 45.60 & \greenthree 28.20 & 43.60   \\
    &GSM8K & 48.45 & \redthree 51.18 & 46.55 & 48.60 & \redtwo 51.25 & \redone 52.46 & 46.47 & \greenone 23.65 & \greentwo 25.09 & 44.96 & \greenthree 26.16 & 46.78   \\
    &Wikinews &\greenthree46.49 &\greentwo45.65  &\greenone32.65 &87.47  &93.76  & 96.41&95.81&\redthree96.46&\redone96.85&90.74&\redtwo96.67&83.47  \\
    \cline{1-14}     \addlinespace[0.2ex]
     \multirow{3}{*}{Llama-3-8B-Instruct}&MBPP& \redtwo 49.60 & \redone 50.00 & 49.00 & \greenthree46.80 & 48.60  & \redthree 49.20 & 49.00 & 48.40 & \greentwo 45.80 & 48.20 & \greenone 45.60 & 48.80   \\
    &GSM8K & 69.60 & \redtwo71.04 & \redthree 69.83 & 68.39 & \redtwo70.43 & 68.61 & 68.76 & \greentwo 65.05 & \greenthree 65.81 & 68.84 & \greenone 64.52 & 69.67   \\
    &Wikinews & 54.13  &\greenthree 51.81  &\greenthree 75.72 &\redone 81.81  &\redtwo 79.41  & 75.28  &68.23  &71.36  &\greenone 37.08  & 72.35  & \greentwo50.53 &\redthree 76.55  \\
    \cline{1-14}     \addlinespace[0.2ex]
     \multirow{3}{*}{vicuna-7b-v1.5}&MBPP& 22.60  &\redtwo 24.60  &\redone 25.80  &21.60  &22.60  &22.40  &\redthree 23.40  &\greenthree 21.20  &\greenone 17.80  &21.60  &\greentwo 19.20  &22.40 \\
    &GSM8K &\greenthree 18.04  &\redone 25.47  &\redtwo 21.68  &\redthree 20.17  &19.03  &19.18  &19.71  &18.95  &\greentwo 16.15  &20.02  &\greenone 15.77&19.71\\
    &Wikinews &\greentwo 84.63  &\redone 94.43  &90.19  &\redthree 90.70  &\redtwo 91.37  &\greenone 80.70  &89.47  &87.65  &85.33  &\greenthree 85.02  &89.58  &89.22 \\
    \addlinespace[-0.5ex]
    \bottomrule
    \end{tabular}
1    }
    \vspace{-0.5em}
    \caption{Results for different foundation models on MBPP, GSM8K, Wikinews datasets. The corresponding hyperparameters for each decoding method are listed in Table~\ref{tab:hyper6}.}
    \vspace{-1.5em}
    \label{tab:model}
\end{table*}
We extend our analysis to investigate the decoding methods under different foundation models\footnote{~CD and DoLa are not included. Because it is challenging to find an amateur model for each foundation model for CD, and for DoLa, it is difficult to determine the appropriate number of layers for logits comparison for individual models.}. We select several representive models, including CodeLlama-7b~\cite{Rozi2023CodeLlama}, Qwen-7B~\cite{qwen}, MPT-7b~\cite{2023mpt}, Mistral-7B~\cite{jiang2023mistral}, deepseek-moe-16b-base~\cite{dai2024deepseekmoe}, Llama-3-8B~\cite{llama3modelcard} and their aligned versions. Addtionally we select vicuna-7b-v1.5~\cite{vicuna2023} which is an SFT-ed model from llama2-7B without RHLF for analysis. It is crucial to underscore that these models vary significantly in several aspects, such as pre-training data, model architecture, and etc. As illustrated in Tables~\ref{tab:model}, the results observed in~\cref{sec:performance} are still applicable to LLMs with different architectures. Detailed as below: i) For unaligned models, deterministic methods generally perform better than stochastic methods on all tasks except open-ended text generation. ii) Aligned models are less dependent on decoding methods than unaligned models. iii) Among stochastic methods, temperature sampling generally performs better, particularly when using unaligned models.
Apart from these consistencies, it is worth noting that different decoding methods may result in different performance rankings for LLMs. For instance, Codellama outperforms Qwen by 7.60\% in the MBPP with top-$k$ sampling, yet lags behind by 2.20\% with $\eta$ sampling. This implies that different models still have varying adaptability to specific decoding methods, suggesting that the selection of decoding strategies should be more meticulously rigorous during the evaluation of LLMs.

\section{Settings of Hyperparameters}
\label{appx:hyperparameter}
The optimal hyperparameters for each decoding method across different datasets and models are listed from Table~\ref{tab:llama-7b-p} to Table~\ref{tab:senti_hyper}.
\begin{table*}[t]
    \small
    \centering
    \resizebox{0.99\textwidth}{!}{\begin{tabular}{c|c|cccccccc|cccccc}
    \toprule
    \addlinespace[-0.0001ex]
    \multirow{2}{*}{\textbf{Model}} & \multirow{2}{*}{\textbf{Dataset}}&\multicolumn{8}{c|}{\textbf{Deterministic Methods}}& \multicolumn{6}{c}{\textbf{Stochastic Methods}} \\
    \cline{4-16}
    \addlinespace[0.2ex]
    &&\textbf{Greedy} & \textbf{BS} & \textbf{DBS} &\textbf{CS} & \textbf{FSD} &\textbf{FSD-d} &\textbf{CD} &\textbf{DoLa} &\textbf{Temp} &\textbf{Top-$p$} &\textbf{Top-$k$} &\textbf{$\eta$} &\textbf{Miro} & \textbf{Typical} \\
    \cline{1-16}
    \addlinespace[0.5ex]
    \cline{2-16}
    \addlinespace[0.2ex]
    \multirow{14}{*}[-0.6cm]{\rotatebox{90}{\large Llama2-7B}}
   
    &HumanEval& - &8 &8\_2 &0.4 &0.2 &0.4 &0.3 &[16,32) &0.4 &0.8 &20 &0.0006 &5.0 &0.95 \\
    &MBPP& - &4 &8\_2 &0.3 &0.2 &0.4 &0.3 &[16,32) &0.3 &0.8 &5 &0.002 &4.0 &0.9\\
    \cline{2-16}
    \addlinespace[0.2ex] 
    &GSM8K& - &8 &4\_2 &0.4 &0.3 &0.4 &0.3 &[0, 16) &0.2 &0.8 &5 &0.004 &5.0 &0.9 \\
    \cline{2-16}
    \addlinespace[0.2ex]
    &XSUM& - &4 &4\_4 &0.1 &0.1 &0.2 &0.1 &[0, 16) &0.2 &0.8 &5 &0.004 &2.5 &0.92 \\
    &CNN/DM& - &4 &4\_4 &0.3 &0.2 &0.3 &0.1 &[0, 16) &0.4 &0.8 &5 &0.002 &2.5 &0.9\\
    \cline{2-16}
    \addlinespace[0.2ex]
    &De$\Rightarrow$En&- &8 &4\_2 &0.1 &0.1 &0.3 &0.1 &[0, 16) &0.1 &0.8 &5 &0.004 &2.5 &0.9\\
    &En$\Rightarrow$De& - &4 &4\_2 &0.1 &0.1 &0.3 &0.1 &[0, 16) &0.1 &0.8 &5 &0.004 &2.5 &0.9 \\
    &Zh$\Rightarrow$En& - &4 &4\_2 &0.2 &0.1 &0.5 &0.1 &[0, 16) &0.1 &0.8 &5 &0.004 &2.5 &0.9 \\
    &En$\Rightarrow$Zh& - &4 &4\_4 &0.2 &0.2 &0.1 &0.1 &[16,32) &0.1 &0.8 &5 &0.004 &2.5 &0.9\\
    \cline{2-16}
    \addlinespace[0.2ex]
    &CQA& - &4 &8\_4 &0.4 &0.2 &0.2 &0.1 &[16,32) &0.2 &0.85 &5 &0.004 &3.0 &0.9 \\
    &SQA& - &4 &8\_2 &0.5 &0.3 &0.3 &0.7 &[0, 16) &0.3 &0.85 &5 &0.0006 &5.0 &0.92 \\
    \cline{2-16}
    \addlinespace[0.5ex] 
    \cline{2-16}
    \addlinespace[0.2ex]
     &Wikinews& - &4 &8\_2 &0.6 &0.5 &0.6 &0.9 &[16,32) &0.8 &0.85 &10 &0.0003 &2.5 &0.92\\
    &Wikitext& - &4 &4\_2 &0.6 &0.4 &0.6 &0.9 &[0, 16) &0.9 &0.8 &20 &0.002 &5.0 &0.95 \\
    &Book& - &4 &4\_2 &0.6 &0.4 &0.5 &0.9 &[0, 16) &0.8 &0.95 &50 &0.0006 &3.0 &0.2\\
    \cline{1-16}
    \addlinespace[0.5ex]
    \cline{2-16}
    \addlinespace[0.2ex] 
    \multirow{12}{*}[-0.3cm]{\rotatebox{90}{\large Llama2-7B-Chat}}
    & HumanEval &
   - &8 &8\_4 &0.5 &0.4 &0.5 &0.9 &[0, 16) &0.1 &0.9 &5 &0.0003 &3.0 &0.9 \\
    &MBPP  & - &8 &8\_2 &0.3 &0.2 &0.2 &0.1 &[0, 16) &0.3 &0.8 &5 &0.002 &4.0 &0.95 \\
    \cline{2-16}
    \addlinespace[0.2ex]
    & GSM8K &  
    - &8 &4\_2 &0.3 &0.4 &0.1 &0.7 &[0, 16) &0.5 &0.8 &10 &0.0009 &5.0 &0.95 \\
    \cline{2-16}
    \addlinespace[0.2ex]
    & XSUM & - &8 &8\_2 &0.3 &0.4 &0.4 &0.1 &[0, 16) &0.5 &0.85 &10 &0.0009 &5.0 &0.2 \\
    & CNN/DM  & 
    - &8 &8\_2 &0.1 &0.1 &0.1 &0.3 &[0, 16) &0.5 &0.85 &5 &0.004 &2.5 &0.92 \\
    \cline{2-16}
    \addlinespace[0.2ex]
    & CQA &- &4 &8\_2 &0.2 &0.5 &0.4 &0.5 &[16,32) &0.5 &0.85 &50 &0.002 &4.0 &0.92 \\
    & SQA & 
    - &4 &8\_2 &0.1 &0.4 &0.5 &0.1 &[0, 16) &0.1 &0.85 &100 &0.0009 &4.0 &0.95 \\
    \cline{2-16}
    \addlinespace[0.5ex] 
    \cline{2-16}
    \addlinespace[0.2ex]
    & Wikinews &
    - &4 &8\_2 &0.4 &0.5 &0.3 &0.9 &[16,32) &0.5 &0.8 &5 &0.002 &5.0 &0.95 \\
    & Wikitext &
    - &4 &8\_2 &0.6 &0.6 &0.2 &0.7 &[16,32) &0.1 &0.85 &20 &0.0009 &4.0 &0.95 \\
    & Book & 
    - &8 &4\_2 &0.4 &0.4 &0.5 &0.7 &[16,32) &0.8 &0.85 &10 &0.002 &3.0 &0.95 \\
    \cline{2-16}
    \addlinespace[0.5ex] 
    \cline{2-16}
    \addlinespace[0.2ex]
    & FActScore &  - &8 &4\_2 &0.1 &0.2 &0.4 &0.9 &[16,32) &0.5 &0.8 &5 &0.0006 &5.0 &0.95 \\
    \cline{2-16}
    \addlinespace[0.2ex]
    & AlpacaEval  & - & 8 & 4\_2 &0.1 & 0.2 & 0.4 & 0.9 &1& 0.5& 0.8  & 5 &0.0006 &5.0 &0.95  \\
    \addlinespace[-0.2ex]
    \bottomrule
    \end{tabular}}
    \vspace{-0.5em}
    \caption{Optimal hyperparameter settings in Table~\ref{tab:llama-7b}.}
    \vspace{-1.5em}
    \label{tab:llama-7b-p}
\end{table*}
\begin{table*}[t]
    \small
    \centering
    \resizebox{0.99\textwidth}{!}{\begin{tabular}{c|c|cccccccc|cccccc}
    \toprule
    \addlinespace[-0.0001ex]
    \multirow{2}{*}{\textbf{Model}} & \multirow{2}{*}{\textbf{Dataset}}& \multicolumn{8}{c|}{\textbf{Deterministic Methods}}& \multicolumn{6}{c}{\textbf{Stochastic Methods}} \\
    \cline{3-16}
    \addlinespace[0.2ex]
    &&\textbf{Greedy} & \textbf{BS} & \textbf{DBS} &\textbf{CS} & \textbf{FSD} &\textbf{FSD-d} &\textbf{CD} &\textbf{DoLa} &\textbf{Temp} &\textbf{Top-$p$} &\textbf{Top-$k$} &\textbf{$\eta$} &\textbf{Miro} & \textbf{Typical}\\
    \cline{1-16}
    \addlinespace[0.2ex]
    \multirow{3}{*}{7B} &MBPP& - &4 &8\_2 &0.3 &0.2 &0.4 &0.3 &[16,32) &0.3 &0.8 &5 &0.002 &4.0 &0.9\\
    &GSM8K& - &8 &4\_2 &0.4 &0.3 &0.4 &0.3 &[0,16) &0.2 &0.8 &5 &0.004 &5.0 &0.9 \\
    &Wikinews& - &4 &8\_2 &0.6 &0.5 &0.6 &0.9 &[16,32) &0.8 &0.85 &10 &0.0003 &2.5 &0.92\\
    \cline{1-16}
    \addlinespace[0.2ex]
    \multirow{3}{*}{13B} &MBPP&- &4 &8\_2 &0.2 &0.3 &0.4 &0.3 &[0,20) &0.3 &0.85 &5 &0.002 &2.5 &0.2\\
    &GSM8K & 
    - &4 &8\_2 &0.2 &0.3 &0.4 &0.3 &[0,20) &0.1 &0.8 &5 &0.002 &2.5 &0.2\\
    &Wikinews & - &4 &4\_2 &0.5 &0.4 &0.3 &0.9 &[0,20) &0.7 &0.95 &50 &0.004 &5.0 &0.9\\
    \cline{1-16}
    \addlinespace[0.2ex]
    \multirow{3}{*}{70B} &MBPP&- &8 &4\_4 &0.6 &0.1 &0.4 &0.3 &[0,20) &0.1 &0.8 &5 &0.0003 &5.0 &0.2\\
    &GSM8K&- &4 &4\_2 &0.2 &0.2 &0.5 &0.9 &[0,20) &0.4 &0.8 &5 &0.0006 &5.0 &0.2\\
    &Wikinews &- &4 &8\_2 &0.6 &0.1 &0.6 &0.9 &[60,80) &0.9 &0.85 &50 &0.002 &3.0 &0.2\\    
    \cline{1-16}
    \addlinespace[0.2ex]
    \multirow{3}{*}{7B-chat} &MBPP& - &4 &8\_2 &0.3 &0.2 &0.4 &0.3 &[16,32) &0.3 &0.8 &5 &0.002 &4.0 &0.9\\
    &GSM8K &- &8 &4\_2 &0.3 &0.4 &0.1 &0.7 &[0,16) &0.5 &0.8 &10 &0.0009 &5.0 &0.95\\
    &Wikinews & - &4 &8\_2 &0.4 &0.5 &0.3 &0.9 &[16,32) &0.5 &0.8 &5 &0.002 &5.0 &0.95\\
    \cline{1-16}
    \addlinespace[0.2ex]
    \multirow{3}{*}{13B-chat} &MBPP& - &8 &8\_2 &0.3 &0.3 &0.4 &0.9 &[20,40) &0.3 &0.95 &5 &0.0003 &4.0 &0.2\\
    &GSM8K&- &8 &8\_2 &0.4 &0.2 &0.5 &0.7 &[20,40) &0.4 &0.9 &50 &0.004 &5.0 &0.92\\
    &Wikinews& - &8 &4\_4 &0.5 &0.4 &0.6 &0.7 &[20,40) &0.5 &0.8 &50 &0.0006 &3.0 &0.9\\
    \cline{1-16}
    \addlinespace[0.2ex]
    \multirow{3}{*}{70B-chat} &MBPP& - &8 &8\_2 &0.6 &0.2 &0.1 &0.9 &[60,80) &0.6 &0.9 &5 &0.0006 &2.5 &0.2 \\
    &GSM8K&- &4 &8\_2 &0.4 &0.2 &0.4 &0.1 &[60,80) &0.3 &0.8 &20 &0.004 &2.5 &0.9\\
    &Wikinews& - &4 &4\_2 &0.4 &0.6 &0.2 &0.9 &[0,20) &0.6 &[60,80) &5 &0.002 &2.5 &0.95\\
    \addlinespace[-0.5ex]
    \bottomrule
    \end{tabular}}
    \caption{Optimal hyperparameter settings in Table~\ref{tab:scale}.}
    \label{tab:hyper4}
\end{table*}
\begin{table*}[t]
    \small
    \centering
    \resizebox{0.99\textwidth}{!}{\begin{tabular}{c|c|cccccccc|cccccc}
    \toprule
    \addlinespace[-0.0001ex]
    \multirow{2}{*}{\textbf{Model}} & \multirow{2}{*}{\textbf{Dataset}}& \multicolumn{8}{c|}{\textbf{Deterministic Methods}}& \multicolumn{6}{c}{\textbf{Stochastic Methods}} \\
    \cline{3-16}
    \addlinespace[0.2ex]
    &&\textbf{Greedy} & \textbf{BS} & \textbf{DBS} &\textbf{CS} & \textbf{FSD} &\textbf{FSD-d} &\textbf{CD} &\textbf{DoLa} &\textbf{Temp} &\textbf{Top-$p$} &\textbf{Top-$k$} &\textbf{$\eta$} &\textbf{Miro} & \textbf{Typical} \\
    \cline{1-16}
    \addlinespace[0.2ex]
    \multirow{3}{*}{13B-INT4}&MBPP&  - &4 &4\_4 &0.1 &0.2 &0.4 &0.3 &[0,20) &0.3 &0.8 &10 &0.004 &4.0 &0.92\\
    &GSM8K &  - &4 &4\_2 &0.3 &0.3 &0.3 &0.3 &[0,20) &0.4 &0.8 &5 &0.0003 &2.5 &0.2\\
    &Wikinews & - &4 &4\_2 &0.3 &0.5 &0.5 &0.9 &[0,20) &0.9 &0.8 &100 &0.0006 &3.0 &0.95\\
    \cline{1-16}     \addlinespace[0.2ex]
    \multirow{3}{*}{13B-INT8} &MBPP& - &4 &8\_4 &0.5 &0.3 &0.4 &0.3 &[0,20) &0.4 &0.8 &5 &0.0006 &4.0 &0.92\\
    &GSM8K & - &4 &4\_2 &0.6 &0.2 &0.5 &0.3 &[0,20) &0.1 &0.8 &5 &0.004 &3.0 &0.9 \\ 
    &Wikinews& - &4 &4\_2 &0.5 &0.2 &0.6 &0.9 &[0,20) &0.9 &0.85 &10 &0.0009 &3.0 &0.95\\
    \cline{1-16}     \addlinespace[0.2ex]
    \multirow{3}{*}{13B-Chat-INT4}&MBPP&- &8 &8\_4 &0.1 &0.3 &0.2 &0.1 &[20,40) &0.1 &0.9 &5 &0.0009 &2.5 &0.2\\
    &GSM8K &  - &8 &8\_2 &0.1 &0.3 &0.5 &0.1 &[20,40) &0.5 &0.85 &10 &0.004 &3.0 &0.92 \\
    &Wikinews &- &8 &4\_2 &0.2 &0.6 &0.6 &0.9 &[20,40) &0.7 &0.95 &5 &0.0006 &2.5 &0.2\\
    \cline{1-16}     \addlinespace[0.2ex]
    \multirow{3}{*}{13B-Chat-INT8} &MBPP&  - &4 &8\_2 &0.4 &0.2 &0.5 &0.3 &[20,40) &0.9 &0.9 &5 &0.0009 &3.0 &0.92\\
    & GSM8K &  - &4 &8\_2 &0.3 &0.3 &0.3 &0.3 &[20,40) &0.5 &0.8 &100 &0.0009 &3.0 &0.95\\
&Wikinews&- &8 &4\_4 &0.5 &0.1 &0.4 &0.7 &[20,40) &0.8 &0.95 &5 &0.002 &3.0 &0.2\\
    \addlinespace[-0.5ex]
    \bottomrule
    \end{tabular}}
    \caption{Optimal hyperparameter settings in Table~\ref{tab:quantization}.}
    \label{tab:hyper5}
\end{table*}
\begin{table*}[t]
    \small
    \centering
    \resizebox{0.99\textwidth}{!}{\begin{tabular}{c|c|cccccc|cccccc}
    \toprule
    \addlinespace[-0.0001ex]
    \multirow{2}{*}{\textbf{Model}} & \multirow{2}{*}{\textbf{Dataset}}& \multicolumn{6}{c|}{\textbf{Deterministic Methods}}& \multicolumn{6}{c}{\textbf{Stochastic Methods}} \\
    \cline{3-14}
    \addlinespace[0.2ex]
    &&\textbf{Greedy} & \textbf{BS} & \textbf{DBS} &\textbf{CS} & \textbf{FSD} &\textbf{FSD-d}  &\textbf{Temp} &\textbf{Top-$p$} &\textbf{Top-$k$} &\textbf{$\eta$} &\textbf{Miro} & \textbf{Typical} \\
    \cline{1-14}
    \addlinespace[0.2ex]
    \multirow{3}{*}{CodeLlama-7b}&MBPP& - &4 &4\_4 &0.3 &0.1 &0.3 &0.6 &0.8 &5 &0.004 &4.0 &0.2 \\
    &GSM8K & - &4 &8\_2 &0.3 &0.2 &0.5 &0.2 &0.8 &5 &0.004 &5.0 &0.95\\
    &Wikinews &- &8 &4\_2 &0.6 &0.6 &0.6 &0.9 &0.9 &50 &0.004 &2.5 &0.2\\
    \cline{1-14}     \addlinespace[0.2ex]
    \multirow{3}{*}{CodeLlama-7b-Instruct}&MBPP& - &4 &8\_4 &0.5 &0.3 &0.2 &0.3 &0.8 &5 &0.002 &2.5 &0.9\\
    &GSM8K &- &4 &8\_2 &0.3 &0.1 &0.4 &0.2 &0.8 &5 &0.0003 &5.0 &0.2\\
    &Wikinews & - &4 &8\_2 &0.3 &0.2 &0.3 &0.8 &1 &10 &0.0003 &2.5 &0.95  \\
    \cline{1-14}     \addlinespace[0.2ex]
    \multirow{3}{*}{Qwen-7B}&MBPP& - &4 &4\_2 &0.3 &0.1 &0.1 &0.1 &0.85 &5 &0.0009 &2.5 &0.2 \\
    &GSM8K &- &4 &8\_2 &0.5 &0.1 &0.5 &0.1 &0.8 &5 &0.002 &2.5 &0.2\\
    &Wikinews & - &4 &8\_4 &0.1 &0.3 &0.3 &0.4 &0.8 &50 &0.0009 &2.5 &0.92 \\
    \cline{1-14}     \addlinespace[0.2ex]
     \multirow{3}{*}{Qwen-7B-Chat}&MBPP&- &4 &8\_4 &0.4 &0.1 &0.1 &0.2 &0.85 &50 &0.004 &3.0 &0.2\\
    &GSM8K & - &8 &8\_4 &0.1 &0.3 &0.2 &0.2 &0.85 &10 &0.0009 &3.0 &0.2 \\
    &Wikinews & - &4 &4\_4 &0.3 &0.3 &0.1 &0.3 &0.95 &10 &0.0006 &3.0 &0.92 \\
    \cline{1-14}     \addlinespace[0.2ex]
    \multirow{3}{*}{MPT-7b}&MBPP& - &8 &8\_2 &0.5 &0.1 &0.3 &0.1 &0.85 &5 &0.004 &3.0 &0.92\\
    &GSM8K & - &4 &4\_2 &0.5 &0.3 &0.5 &0.2 &0.9 &5 &0.004 &5.0 &0.9 \\
    &Wikinews &  - &4 &8\_2 &0.6 &0.4 &0.5 &0.8 &0.8 &50 &0.0006 &5.0 &0.95 \\
    \cline{1-14}     \addlinespace[0.2ex]
    \multirow{3}{*}{MPT-7b-Instruct}&MBPP&- &8 &8\_2 &0.4 &0.3 &0.4 &0.2 &0.8 &5 &0.0009 &4.0 &0.95 \\
    &GSM8K & - &4 &4\_4 &0.2 &0.3 &0.4 &0.5 &0.9 &10 &0.0006 &2.5 &0.2 \\
    &Wikinews & - &4 &8\_4 &0.3 &0.3 &0.5 &0.1 &0.85 &50 &0.002 &4.0 &0.2 \\
    \cline{1-14}     \addlinespace[0.2ex]
    \multirow{3}{*}{Mistral-7B}&MBPP& - & 4  & 4\_4& 0.1& 0.1& 0.4   & 0.9 & 0.95 & 20 & 0.004 & 2.5 & 0.95 \\
    &GSM8K & -  & 8 & 8\_2& 0.1& 0.6& 0.2 & 0.4 & 0.85 & 20 & 0.002 & 2.5 & 0.2     \\
    &Wikinews &  -&4&4\_2&0.5&0.5&0.4&0.8&0.9&20&0.0009&5.0& 0.95\\
    \cline{1-14}     \addlinespace[0.2ex]
    \multirow{3}{*}{Mistral-7B-Instruct}&MBPP&- & 4  & 8\_4& 0.4& 0.5& 0.2   & 0.4  & 0.85  & 100   & 0.004  & 2.5  & 0.95 \\
    &GSM8K & - & 4  & 8\_2& 0.3& 0.1& 0.6   & 0.5  & 0.8   & 10    & 0.0009& 2.5  & 0.2 \\
    &Wikinews & -& 4&8\_4&0.5&0.2&0.1&0.7&0.95&5&0.0003&3.0&0.9 \\
    \cline{1-14}     \addlinespace[0.2ex]
    \multirow{3}{*}{deepseek-moe-16b-base}&MBPP& - & 4 &4\_2                 & 0.4 & 0.3 & 0.1 & 0.3 & 0.9 & 50  & 0.0003               & 5.0 & 0.9 \\
    &GSM8K & - & 4 & 4\_2                 & 0.5 & 0.3 & 0.2 & 0.4 & 1   & 10  & 0.004                & 3.0 & 0.95                 \\
    &Wikinews &  - &4 &4\_2 &0.4&0.2&0.5&0.7&0.9&10&0.0003&3&0.9 \\
    \cline{1-14}     \addlinespace[0.2ex]
    \multirow{3}{*}{deepseek-moe-16b-chat}&MBPP&- & 4 & 8\_2                 & 0.1 & 0.4 & 0.2 & 0.6 & 0.8 & 5   & 0.002                & 4.0 & 0.92                 \\
    &GSM8K & - & 4 & 8\_2                 & 0.2 & 0.3 & 0.1 & 0.4 & 0.8 & 5   & 0.004                & 5.0 & 0.9 \\
    &Wikinews & - &8 &4\_4 &0.6 &0.4 &0.5 &0.9 &0.85 &50 &0.0006 &5 &0.95 \\
    
    \cline{1-14}
    \addlinespace[0.2ex]
    \multirow{3}{*}{Llama-3-8B}&MBPP& - & 4 & 8\_2& 0.3& 0.2& 0.3  & 0.9 & 1    & 20   & 0.002 & 3.0 & 0.2 \\
    &GSM8K & - & 4 & 8\_2& 0.2& 0.1& 0.5  & 0.4 & 1    & 50   & 0.0009& 4.0 & 0.92    \\
    &Wikinews &  - &8 &8\_2 &0.3&0.3&0.5&0.8&0.8&100&0.0003&3&0.2 \\
    \cline{1-14}     \addlinespace[0.2ex]
    \multirow{3}{*}{Llama-3-8B-Instruct}&MBPP&- & 8 & 4\_4& 0.6& 0.6& 0.3  & 0.2 & 0.8  & 20   & 0.0006& 2.5 & 0.2 \\
    &GSM8K & - & 8 & 4\_2& 0.1& 0.4& 0.3  & 0.1 & 1    & 5    & 0.002 & 3.0 & 0.95 \\
    &Wikinews & - &8 &4\_4 &0.3 &0.1 &0.6 &0.2 &0.8 &100 &0.0006 &5 &0.95 \\
    \cline{1-14}     \addlinespace[0.2ex]
    \multirow{3}{*}{vicuna-7b-v1.5}&MBPP&- &8 &8\_2 &0.1 &0.4 &0.4 &0.3 &0.85 &5 &0.0006 &5.0 &0.9 \\
    &GSM8K &-&8 &8\_2 &0.2 &0.1 &0.1 &0.4 &0.8 &5 &0.0009 &2.5 &0.2 \\
    &Wikinews &  - &8 &4\_4 &0.6 &0.6 &0.2 &0.5 &0.9 &10 &0.0009 &3.0 &0.95 \\
    \addlinespace[-0.5ex]
    \bottomrule
    \end{tabular}}
    \caption{Optimal hyperparameter settings in Table~\ref{tab:model}.}
    \label{tab:hyper6}
\end{table*}
\begin{table*}[t]
    \small
    \centering
    \resizebox{0.99\textwidth}{!}{
    \begin{tabular}{c|l|cccccccc|cccccc}
    \toprule
    \addlinespace[-0.0001ex]
    &  & \multicolumn{8}{c|}{\textbf{Deterministic Methods}} & \multicolumn{6}{c}{\textbf{Stochastic Methods}} \\
    \cline{3-16}
    \addlinespace[0.2ex]
    \multirow{-2}{*}{\textbf{Model}} & \multirow{-2}{*}{\textbf{Setting}} & \textbf{Greedy} & \textbf{BS} & \textbf{DBS} &\textbf{CS} & \textbf{FSD} &\textbf{FSD-d} &\textbf{CD} &\textbf{DoLa} &\textbf{Temp} &\textbf{Top-$p$} &\textbf{Top-$k$} &\textbf{$\eta$} &\textbf{Miro} & \textbf{Typical} \\
    \cline{1-16}
    \addlinespace[0.2ex] 
    & $\text{Score}_{\text{best}}$ & 78.47 & 80.97 & 79.26 & 94.49 & \redtwo 96.71 & \redone 97.52 & 92.90 & 89.94 & \redthree 95.36 & 80.61 & 74.70 & 71.44 & 67.68 & 75.68 \\
    & $\text{Score}_{\lambda}$ & 78.47 & 80.68 & 76.53 & \redthree 91.85 & \redtwo 94.32 & \redone 94.76 & 83.48 & 86.34 & 83.77 & 79.65 & 71.80 & 69.61 & 64.93 & 74.15 \\
    & Param. & - & 4 & 4\_2 & 0.4 & 0.2 & 0.4 & 0.3 & 0 & 0.6 & 0.8 & 0.5 & 0.004 & 2.5 & 0.9 \\
    \multirow{-4}{*}{Llama2-7B} & Drop & 0.00 & 0.29 & 2.74 & 2.64 & 2.39 & 2.76 & 9.42 & 3.60 & 11.59 & 0.96 & 2.89 & 1.83 & 2.76 & 1.53 \\
    \cline{1-16}
    \addlinespace[0.2ex]
    & $\text{Score}_{\text{best}}$ & 87.90 & \redone 96.39 & \redtwo 95.79 & 91.61 & 94.97 & 93.50 & 92.75 & 74.99 & \redthree 94.97 & 91.83 & 91.82 & 91.45 & 88.71 & 92.41  \\
    & $\text{Score}_{\lambda}$ & 87.90 & \redtwo 95.03 & \redone 95.54 & 89.70 & \redthree 91.37 & 91.02 & 89.40 & 73.32 & 91.06 & 89.57 & 89.41 & 88.88 & 86.29 & 90.91  \\
    & Param. & - & 8 & 8\_2 & 0.2 & 0.3 & 0.3 & 0.7 & 0 & 0.4 & 0.85 & 0.5 & 0.002 & 4 & 0.95 \\
    \multirow{-4}{*}{Llama2-7B-Chat} & Drop & 0.00 & 1.63 & 0.10 & 2.09 & 4.22 & 2.68 & 3.76 & 1.69 & 4.68 & 2.41 & 2.89 & 2.78 & 2.36 & 1.80 \\
    \bottomrule
    \addlinespace[-0.5ex]
    \end{tabular}}
    \vspace{-0.5em}
    \caption{Hyperparameter Sensitivity. $\text{Score}_{\text{best}}$ and the best $\text{Score}_{\lambda}$ with their optimal hyperparameters on Llama2-7B and Llama2-7B-Chat.
    }
    \vspace{-1.5em}
    \label{tab:senti_hyper}
\end{table*}

\section{Analyses of Generation Diversity}
\begin{table*}[t]
    \small
    \centering
    \resizebox{0.99\textwidth}{!}{\begin{tabular}{c|c|cccccccc|cccccc}
    \toprule
    \addlinespace[-0.0001ex]
    \multirow{2}{*}{\textbf{Model}} & \multirow{2}{*}{\textbf{Dataset}}& \multicolumn{8}{c|}{\textbf{Deterministic Methods}}& \multicolumn{6}{c}{\textbf{Stochastic Methods}}\\
    \cline{3-16}
    \addlinespace[0.2ex]
    &&\textbf{Greedy} & \textbf{BS} & \textbf{DBS} &\textbf{CS} & \textbf{FSD} &\textbf{FSD-d} &\textbf{CD} &\textbf{DoLa} &\textbf{Temp} &\textbf{Top-$p$} &\textbf{Top-$k$} &\textbf{$\eta$} &\textbf{Miro} & \textbf{Typical}\\
    \cline{1-16}
    \addlinespace[0.2ex]
    Llama2-7B & wikinews & \greentwo 1.8& \greenthree 2.9& \greenone 1.5&\redtwo 94.0&\redone 98.9&90.5&43.7&51.7&79.9&83.2&80.8&\redthree 92.3&91.9&90.9 \\
    Llama2-7B-Chat& wikinews &87.7&\greenthree87.2&\greentwo85.9&\redthree90.0&\redtwo93.0&88.8&\redtwo93.2&\greenone47.2&\greenthree87.2&89.1&88.7&87.8&89.0&88.6 \\
    \addlinespace[-0.5ex]
    \bottomrule
    \end{tabular}}
    \vspace{-0.5em}
    \caption{Results for diversity score for Llama2 7B family on wikinews.}
    \vspace{-1.5em}
    \label{tab:diversity}
\end{table*}
Diversity is a more meaningful metric for open-ended tasks than closed-ended ones. Therefore, we report the diversity scores on Wikinews using Llama2-Chat-7B and Llama2-7B models in Table~\ref{tab:diversity}. Specifically, we adopt the diversity measure defined in~\citealp{yang2023frustratingly}, which computes the degree of repetition across all generations at different n-gram levels. It can be observed that best-performing stochastic methods do not necessarily exhibit higher diversity than best-performing deterministic methods. Concretely, for Llama2-7B, the diversity score of FSD is the highest, while for Llama2-Chat, CD obtains the highest.

\section{Analyses of COMET Score on WMT tasks.}

\begin{table*}[t]
    \small
    \centering
    \resizebox{0.99\textwidth}{!}{\begin{tabular}{c|c|cccccccc|cccccc}
    \toprule
    \addlinespace[-0.0001ex]
    \multirow{2}{*}{\textbf{Model}} & \multirow{2}{*}{\textbf{Dataset}}& \multicolumn{8}{c|}{\textbf{Deterministic Methods}}& \multicolumn{6}{c}{\textbf{Stochastic Methods}}\\
    \cline{3-16}
    \addlinespace[0.2ex]
    &&\textbf{Greedy} & \textbf{BS} & \textbf{DBS} &\textbf{CS} & \textbf{FSD} &\textbf{FSD-d} &\textbf{CD} &\textbf{DoLa} &\textbf{Temp} &\textbf{Top-$p$} &\textbf{Top-$k$} &\textbf{$\eta$} &\textbf{Miro} & \textbf{Typical}\\
    \cline{1-16}
    \addlinespace[0.2ex]
    Llama2-7B&de2en&\redthree83.0&\redone83.3&82.8&82.9&\redthree83.0&\redthree83.0&\redtwo83.1&82.4&82.9&80.9&\greenthree79.9&\greenthree78.9&\greenone78.2&\greentwo79.6\\
    Llama2-7B&en2de&80.7&\redone82.5&\redtwo81.7&\redthree80.9&80.8&80.8&80.8&80.1&80.6&74.5&75.0&\greentwo69.9&\greenone67.6&\greenthree70.0\\
    Llama2-7B&en2zh&79.6&\redone81.6&\redtwo80.6&80.1&\redthree80.4&\redthree80.4&79.8&79.7&79.5&73.2&73.5&\greentwo68.0&\greenone64.8&\greenthree69.6\\
    Llama2-7B&zh2en&77.5&77.2&76.9&77.4&\redone78.2&\redtwo78.1&\redthree77.6&\redthree77.6&77.5&\greenthree73.8&74.1&\greentwo71.9&\greentwo71.9&\greenone71.8\\
    \addlinespace[-0.5ex]
    \bottomrule
    \end{tabular}}
    \vspace{-0.5em}
    \caption{Results for COMET score for Llama2 7B on WMT.}
    \vspace{-1.5em}
    \label{tab:COMET}
\end{table*}
Both COMET~\cite{rei-etal-2022-searching} and BLEU~\cite{post2018SacreBLEU} are important metrics for translation tasks. We provide the COMET results for the translation tasks in Table~\ref{tab:COMET}. It can be observed that, similarly, for unaligned models, deterministic methods generally perform better than stochastic methods according to the COMET metric.

\section{Ouput of DoLa}
The output examples that DoLa fails to terminate its generation appropriately are listed in Table~\ref{tab:faileddola}.

\label{appx:dola}
\begin{table*}[h]
    \centering
    \small
    \begin{tabular}{p{0.95\linewidth}}
        \toprule
        \underline{\textbf{\textsc{GSM8K}}} \\
        \vspace{-2mm}
        \textbf{Instruction:} Please answer the math questions below.\\Luke is buying fabric for new curtains. There are five windows. Each window is 35 inches wide and Luke needs to buy fabric equal to 2 times the total width of the windows. How much fabric should he buy?\\You need to first take step-by-step reasoning and then give the final result.\\
        \vspace{-1mm}
        \textbf{Response:} Sure, I'd be happy to help you with that! Here's the step-by-step reasoning:\\1. The total width of the five windows is 5 x 35 = 175 inches.\\2. Luke needs to buy fabric equal to 2 times the total width of the windows, so he needs to buy fabric equal to 2 x 175 = 350 inches.\\Therefore, Luke should buy 350 inches of fabric.\\I hope this helps! Let me know if you have any other questions.b-1-1-1-1-1-1-1-1-1-1-1-1-1-1-1-1-1-1-1-1-1-1-1-1-1-1-1-1-1-1-1-1-1-1-1-1-1-1-1-1-1-1-1-1-1-1-1-1-1-1-1-1-1-1-1-1-1-1-1-1-1-1-1-1-1-1-1-1-1-1-1-1-1-1-1-1-1-1-1-1-1-1-1-1-1-1-1-1-1-1-1-1-1-1-1-1-1-1-1-1-1-1-1-1-1-1-1-1-1-1-1-1-1-1-1-1-1-1-1-1-1-1-1-1-1-1-1-1-1-1-1-1-1-1-1-1-1-1-1-1-1-1-1-1-1-1-1-1-1-1-1-1-1-1-1-1-1-1-1-1-1-1-1-1-1-1-1-1-1-1-1-1-1-1-1-1-1-1-1-1-1-1-1-1-1-1-1-1 \\
        \vspace{2mm}
        \underline{\textbf{\textsc{XSUM}}} \\
        \vspace{-2mm}
        \textbf{Instruction:} Article: The 32-year-old had been playing in the Isthmian Premier Division with Leatherhead following his release by Newport at the end of last season.\\Pidgeley has made 260 appearances in spells with nine clubs, including Chelsea, Watford and Millwall.\\Forest Green are currently second in the National League table, one point behind leaders Cheltenham Town.\\Pidgeley could make his Rovers debut when they host Aldershot on Friday.\\\\Summarize the above article in 1 sentence.\\ \vspace{-1mm}
        \textbf{Response:} Former Newport player, James Pidgeley, who has played for 9 different clubs including Chelsea, Watford, and Millwall, has signed with Forest Green Rovers in the National League and could make his debut against Aldershot on Friday. (No need for a question)oundsMS)) ) ) )) )) )) )) )) )) )) )) )) )) )) )) )) )) )) )) )) )) )) )) )) )) )) )) )) )) )) )) )) )) )) )) )) )) )) )) )) )) )) )) )) )) )) )) )) )) )) )) )) )) )) )) )) )) )) )) )) )) )) )) )) )) )) )) )) )) )) )) )) )) )) )) )) )) )) )) )) )) )) )) )) )) )) )) )) )) )) )) )) )) )) )) )) )) )) )) )) )) )) )) )) )) )) )) )) )) )) )) )) )) )) )) )) )) )) )) )) )) )) )) )) )) )) )) )) )) )) )) )) )) )) )) )) )) )) )) )) )) )) )) )) )) )) )) )) )) )) )) )) )) )) )) )) )) )) )) )) )) )) )) )) )) )) )) )) )) )) )) )) )) )) )) )) )) )) )) )) )) )) )) )) )) )) )) )) )) )) )) )) )) )) )) )) )) )) )) )) )) )) )) )) )) )) )) )) )) )) )) )) )) )) ))\\
        \vspace{2mm}
        \underline{\textbf{\textsc{Wikinews}}} \\
        \vspace{-2mm}
        \textbf{Instruction:} Please help me complete the text continuation based on the following content.
        \\\\
        Snooker player Alex Higgins, nicknamed "Hurricane", was found dead in his apartment in Donegall Road, Belfast, Northern Ireland on Saturday. He was \\
        \vspace{-1mm}
        \textbf{Response:} Sure, here is a possible continuation of the text:\\Snooker player Alex Higgins, nicknamed "Hurricane", was found dead in his apartment in Donegall Road, Belfast, Northern Ireland on Saturday. He was 61 years old and had been struggling with health problems in recent years. The cause of death is currently unknown, but police are treating it as a suspicious circumstances. Higgins was one of the most iconic figures in snooker history, known for his flamboyant style and fierce competitiveness on the table. He won the World Snooker Championship four times and was inducted into the World Snooker Hall of Fame in 2004. His death has sent shockwaves through the snooker community, with many players and fans paying tribute to his legacy.b\\Please let me know if you would like me to make any changes or add anything else. to continue the text. To Complete the Text... To Complete the Text..\\I hope this helps! Let me know if you have any other questions or requests.. MS. MS. MS. MS. MS. MS. MS. MS. MS. MS. MS. MS. MS. MS. MS. MS. MS. MS. MS. MS. MS. MS. MS. MS. MS. MS. MS. MS. MS. MS. MS. MS. MS. MS. MS. MS. MS. MS. MS. MS. MS. MS. MS. MS. MS. MS. MS. MS. MS. MS. MS. MS. MS. MS. MS. MS. MS. MS. MS. MS. MS. MS. MS. MS. MS. MS. MS. MS. MS. MS. MS. MS. MS. MS. MS. MS. MS. MS. MS. MS. MS. MS. MS. MS. MS. MS. MS. MS. MS. MS. MS. MS. MS. MS. MS. MS. MS. MS. MS. MS. MS. MS. MS. MS. MS. MS. MS. MS. MS. MS. MS. MS. MS. MS. MS. MS. MS. MS. MS. MS. MS. MS. MS.\\
        \bottomrule
    \end{tabular}
    \caption{    
    Failed Examples of DoLa.
    }
    \label{tab:faileddola}
\end{table*}

\section{Practical Guidelines}
\label{takeaway}
Our study underscores the significance of selecting an appropriate decoding method in the era of large language models (LLMs). Despite the advancements in LLMs, our findings indicate that the choice of decoding method remains critical and cannot be overlooked. This decision is contingent upon several factors, including the specific test task, the model being used, and the priority—whether it is performance, robustness, or speed. The core contribution of our paper lies in demonstrating the nuanced and complex nature of decoding method selection. The optimal decoding strategy is not universally applicable and varies based on the aforementioned factors. This complexity underscores the necessity for a comprehensive evaluation framework in future research and highlights the need for practitioners to consider multiple dimensions when deploying LLMs. Despite the intricacies involved, we offer several practical guidelines for deploying LLMs without extensive hyperparameter searching: For quick setup, Unaligned Models (e.g., Llama2-7B): For these models, we recommend using either FSD or FSD-d; Aligned Models (e.g., Llama2-7B-Chat): BS or DBS is advised for aligned models to achieve satisfactory performance. When computational resources allow for self-consistency: Unaligned Models (e.g., Llama2-7B): implementing temperature sampling with an optimal temperature setting of 0.7 is recommended to enhance model performance. Aligned Models (e.g., Llama2-7B-Chat): a higher optimal temperature of 0.9 is suggested.
\section{Ethics and Societal Impact}
\paragraph{Ethical Considerations.} Our work highlights the importance of transparency in LLMs, particularly in how decoding methods influence LLM outputs. The variability in performance across tasks and models underscores the need for clear communication about the limitations and potential biases of these systems. Researchers and practitioners must be mindful that the choice of decoding method can significantly impact the generated content, potentially amplifying or mitigating biases present in the underlying models.
\paragraph{Societal Impact.} The findings of this study have far-reaching implications for the deployment of LLMs in real-world applications. By elucidating the trade-offs between performance, robustness, and speed, our work empowers developers to make more informed decisions when implementing these models in diverse contexts. This could lead to more reliable and efficient AI systems in critical areas such as healthcare, education, and public services. However, it also raises concerns about the potential for misuse or overreliance on these systems without a full understanding of their limitations.
The observed task-dependency of decoding methods' performance suggests that careful consideration is needed when applying LLMs to different domains. This is particularly crucial in high-stakes applications where the consequences of model outputs can be significant. Our work also highlights the potential for advanced decoding methods to improve model performance, which could accelerate the adoption of AI technologies across various sectors of society.
\section{Future Work} 
\paragraph{Holistic Evaluations Across Diverse Contexts.} While our study sheds light on the performance, robustness, and speed of various decoding methods, expanding these evaluations to encompass even more varied tasks, languages, and dataset types would provide deeper insights into the generalizability of our findings. This includes tasks like temporal knowledge knowledge graph completion~\cite{luo2024chain}, text-to-sql~\cite{luo2024ptd}, and etc. Additionally, testing with low-resource languages and under-represented dialects is also worth exploring~\cite{zhang2023m3exam}.
\paragraph{User-Centric Evaluation Metrics.} There is a need for developing new evaluation metrics that more directly reflect user satisfaction and real-world efficacy. Incorporating user feedback loops and live deployment scenarios can aid in better understanding the practical utility of different decoding method~\cite{mirowski2023co}.
\paragraph{Extending to New Tasks.} Although our study validates decoding methods across a wide range of tasks, the rapid evolution of LLMs introduces new tasks for future validation. For instance, evaluating models on attributes like honesty and exploring how different methods can contribute to deploying more honest and transparent models is a pertinent area~\cite{li2024survey,zhang2024toolbehonest}. In open-ended scenarios such as human-AI collaboration, beyond simple news generation from a prefix, LLMs need to better cooperate in creative processes to generate both reliable and diverse texts like screenwriting~\cite{chen2024hollmwood}. Future decoding research should thus focus on facilitating such cooperation.
\paragraph{Extending to Large Multimodal Models.} While our current focus is on decoding for LLMs, future work should extend to examining decoding methods in large multimodal models~\cite{openai2024gpt4o,chen2023internvl}. Investigating the effectiveness of these methods in text-to-image, multimodal question answering~\cite{wang2024charxiv}, math reasoing~\cite{lu2023mathvista} and code generation~\cite{shi2024chartmimic} necessitates the attention of future work.

\end{document}